\documentclass[11pt,letter]{article}
\usepackage[english]{babel}
\usepackage{amssymb,amsmath,graphicx}
\usepackage{amsthm}
\usepackage{float}
\usepackage{afterpage}
\usepackage{color}
\usepackage[round,authoryear]{natbib}
\usepackage{enumitem}
\usepackage{setspace}
\usepackage{algorithmic}
\usepackage{algorithm}
\usepackage[table]{xcolor}
\usepackage{multirow}
\usepackage{microtype}
\usepackage{tcolorbox}
\usepackage{bbold}
\usepackage{array}
\usepackage{rotating}
\usepackage{url}
\usepackage{booktabs}
\usepackage{mathtools}
 
\usepackage{natbib}
 \bibpunct[, ]{(}{)}{,}{a}{}{,}%
 \def\newblock{\ }%

\tcbuselibrary{skins} 
\tcbset{
  ycornercontinuous/.style={
    enhanced,
    overlay={
      \node[
        anchor=east,
        fill=none,
        draw=none,
        inner sep=0pt,
        font=\small,
        text=white
      ] at ([xshift=-1.5mm]title.east)
      {$\mathcal{Y}=\mathbb{R}$};
    }
  }
}
\tcbset{ 
  ycornerbinary/.style={
    enhanced,
    overlay={
     \node[
        anchor=east,
        fill=none,
        draw=none,
        inner sep=0pt,
        font=\small,
        text=white
      ] at ([xshift=-1.5mm]title.east)
      {$\mathcal{Y}=\{-1,+1\}$};
    }
  }
}

\tcbset{ 
  ycornermulticlass/.style={
    enhanced,
    overlay={
     \node[
        anchor=east,
        fill=none,
        draw=none,
        inner sep=0pt,
        font=\small,
        text=white
      ] at ([xshift=-1.5mm]title.east)
      {$\lvert \mathcal{Y} \rvert \geq 2$};
    }
  }
}
\newcounter{highlight}
\renewcommand{\thehighlight}{\arabic{highlight}}
\newtcolorbox{highlight}[2][]{colback=white, colframe=gray, boxrule=0.5mm, sharp corners, top=-2pt, title=Highlight \thehighlight. #2, #1}

\title{Trustworthy Machine Learning through the Lens of Combinatorial Optimization:  Survey and Research Perspectives}

\author{Thibaut Vidal}

\begin{document}

\begin{center}

\vspace*{0.5cm}

\begin{LARGE}
Trustworthy Machine Learning through the Lens of Combinatorial \vspace*{0.2cm} \\ Optimization: Survey and Research Perspectives
\end{LARGE}

\vspace*{1cm}

\textbf{Thibaut Vidal, Julien Ferry} \\
CIRRELT \& SCALE-AI Chair in Data-Driven Supply Chains, Department of Mathematics and Industrial Engineering, Polytechnique Montreal, Montreal, Canada\\
thibaut.vidal@polymtl.ca \\

\vspace*{2cm}

\end{center}

\noindent
\textbf{Abstract.}
Modern machine learning (ML) increasingly relies on complex models whose behavior is difficult to characterize beyond empirical performance metrics. Across a wide range of tasks, including prediction, generation, and decision-making, models with similar empirical performance can exhibit markedly different properties in terms of their transparency, interpretability, robustness, fairness, privacy, and certifiability. This survey highlights how optimization- and certification-oriented reasoning can provide a useful framework for reasoning about such differences, supporting tasks ranging from model training and selection to auditing and certification. We review and synthesize recent advances at the intersection of combinatorial optimization (CO) and trustworthy ML, covering both training and post-training tasks, including interpretable model learning, explanation generation, robustness analysis, fairness auditing, model compression, and privacy attacks and protections. Across these domains, CO formulations offer additional capabilities over purely heuristic approaches, e.g., gradient-based ones, notably global guarantees, formal certificates, and explicit treatment of trade-offs. While scalability remains an important challenge, continued progress in solvers and hybrid algorithms suggests a growing role for CO in the design and deployment of trustworthy ML systems.


\vspace*{0.2cm}


\vspace*{0.5cm}

\thispagestyle{empty}
\pagenumbering{arabic}
\newpage

\section{Introduction}
\label{sec:introduction}

Machine learning (ML) has witnessed unprecedented progress over the past decade, rapidly becoming ubiquitous across domains where prediction, pattern recognition, content generation, and automated decision-making increasingly match or surpass human performance. Modern ML systems now underpin applications ranging from computer vision and natural language processing to credit scoring, medical diagnosis, recommender systems, supply chain optimization, transportation, and autonomous systems. In many of these settings, ML models no longer merely assist human judgment but directly inform or automate high-stakes decisions.

At the same time, the deployment of ML in consequential contexts has made it increasingly clear that predictive accuracy alone is insufficient. A growing number of incidents have revealed systematic shortcomings in the design, evaluation, and deployment of ML systems. These include biased or arbitrary decisions (e.g., recidivism prediction tools that are opaque and no more accurate or fair than non-expert human judgment; \citealt{Dressel2018}; or recruitment tools found to discriminate on the basis of age; \citealt{EEOC2022}), security and reliability failures (e.g., coding assistants causing catastrophic data loss; \citealt{Nolan2025}), and, in extreme cases, harmful interactions with users (language models assisting suicide or encouraging delusional behavior; \citealt{Hill2025a,Hill2025}). These risks are further amplified as decision and action capabilities are increasingly delegated to autonomous agents operating with limited human oversight.

All these examples point in the same direction: the responsible use of ML requires methodological tools that go well beyond training or fine-tuning models on available data. Ensuring that ML systems are trustworthy (transparent, explainable, robust, fair, and privacy-preserving) has therefore emerged as a formidable challenge for the coming decades. Importantly, technical approaches to trustworthiness are not optional refinements or post hoc fixes; they fundamentally shape how models are designed, evaluated, selected, and deployed in practice, and must be considered alongside broader AI governance and institutional measures \citep{Reuel2024}.

An important observation motivating this integrative survey is that, for many real-world predictive tasks, there is rarely a single “best” model. Instead, multiple models often achieve near-identical predictive performance while relying on different features, structures, or inductive biases. This phenomenon, famously dubbed the \emph{Rashomon effect} by \citet{Breiman2001a}, implies that model choice cannot be justified by accuracy alone. Rather than being a nuisance, the Rashomon effect creates opportunities: when many models perform similarly, one can select models that satisfy additional trustworthiness desiderata without sacrificing performance \citep{Semenova2022}. In parallel, Occam’s razor suggests that, all else being equal, simpler and more structured models should be preferred, often yielding superior interpretability and reliability. Together, these observations shift the central question from ``which model is most accurate?'' to ``which accurate model should we choose?''

Combinatorial optimization (CO) provides a natural yet historically underutilized framework for addressing this question. CO concerns feasibility and optimization problems over structured, discrete, or combinatorial solution spaces, expressed by explicit objective functions and constraints. While ML and CO have long evolved along largely separate trajectories, a growing body of work has demonstrated the use of CO to train interpretable models, such as optimal decision trees, rule lists, or sparse linear predictors. What is new, and central to this survey, is the rapid expansion of CO and formal verification techniques beyond training, toward a much broader range of \emph{post-training} and \emph{audit} tasks that lie at the core of trustworthy ML.

Recent advances show that many trustworthiness questions can be naturally cast as optimization or feasibility problems: verifying robustness against adversarial perturbations, certifying fairness constraints, generating minimal, plausible, and actionable explanations, auditing privacy leakage through reconstruction attacks, compressing models without altering their behavior, or systematically exploring Rashomon sets of good models. In these settings, CO brings capabilities that are difficult to replicate with purely heuristic or gradient-based methods: global optimality, formal certificates, explicit trade-off analysis, and the ability to integrate rich structural and domain constraints. At the same time, these benefits come with well-known scalability challenges, making careful modeling choices and algorithm design essential.

This survey provides an overarching synthesis of the rapidly growing literature on the application of CO techniques to trustworthy ML. Our goal is to unify a fragmented body of work spanning operations research, theoretical computer science, machine learning, and formal methods. In earlier surveys on ML and optimization, \citet{Gambella2020} primarily focused on training tasks common at the time (e.g., regression, classification, clustering), while \citet{Bengio2021c} examined how ML can be used to improve CO algorithms. More recently, \citet{Justin2025} presented a tutorial treatment of mixed-integer programming (MIP) for responsible ML, focusing primarily on learning tasks. In contrast, our survey adopts a much broader perspective: we cover both training tasks and post-training tasks, span multiple CO paradigms beyond MIP (including SAT, SMT, CP, MaxSAT, and B\&B hybrids), and place particular emphasis on verification, explanation, model simplification, fairness, robustness, and privacy audit.

\paragraph{Survey methodology.}
The scope of this survey is necessarily broad and interdisciplinary, precluding a fully systematic review. We therefore adopt a structured, subdomain-driven approach. Starting from general queries that combine terms such as \emph{machine learning} with \emph{integer programming} or \emph{combinatorial optimization}, we refined the search using keywords specific to each trustworthiness dimension (transparency, interpretability, explainability, robustness, fairness, and privacy). Highly cited works and surveys were used as anchors to identify additional relevant contributions through backward and forward citation exploration. The final selection emphasizes papers published in leading journals and conferences, complemented by recent preprints presenting other key methodological advances. Throughout the survey, we prioritize representativeness over exhaustiveness. In addition, to illustrate the breadth and diversity of CO techniques discussed in the paper, we present compact, self-contained mathematical formulations for representative tasks in a small set of selected \emph{highlights}. The notation used throughout these formulations is adapted from the original papers for consistency and is summarized in Table~\ref{tab:notations_summary} in Appendix~\ref{appendix:notations}.

\paragraph{Contributions.}
This survey makes the following contributions:
(i)   a unifying methodological synthesis of trustworthy ML through the lens of combinatorial optimization;
(ii)  a structured organization of diverse trustworthiness tasks (training, explanation, robustness, fairness, verification, and privacy), highlighting their connections and specificities;
(iii) a curated collection of representative mathematical formulations and algorithmic strategies across multiple CO paradigms; and
(iv)  a timely discussion of current limitations, trade-offs, and emerging research directions at this intersection.

\paragraph{Structure.}
The remainder of the paper is organized around the main desiderata of trustworthy machine learning. For each, we identify key problems that admit CO-related formulations and review corresponding solution approaches. Selected models are highlighted and discussed in greater detail to illustrate the diversity of formulations and algorithms. Section~2 reviews core concepts in trustworthy ML and CO. Sections~3 to 7, respectively, cover transparency, explainability, fairness, robustness, and privacy through CO lenses. Finally, Section~8 concludes with a discussion on important open challenges and research perspectives.

\section{Background}
\label{sec:Defs}

\paragraph{Trustworthy Machine Learning.}
Experience with real-world ML systems has repeatedly shown that predictive accuracy alone is insufficient for responsible and reliable deployment. \emph{Trustworthy ML} therefore refers to a collection of additional desiderata that guide the design, evaluation, and deployment of models in practice \citep{Li2023a}. While definitions vary across communities and application domains, several recurring dimensions are central to this survey:
\begin{itemize}
\item \textbf{Transparency and interpretability} concern the extent to which humans can directly understand a model’s structure and prediction process. Transparency typically assumes full (i.e., ``white-box'') access to the model and emphasizes models whose behavior can be inspected, reasoned about, and audited without complex auxiliary tools. Examples include decision trees, rule lists, scoring systems, and generalized additive models. Transparency and interpretability by design facilitate debugging, accountability, and error detection, and make it easier to identify failure modes and unintended behaviors. As discussed in \citet{Rudin2024}: ``understandable models have understandable flaws.''

\item \textbf{Explainability} refers to the use of algorithmic techniques to provide explanations for a model’s predictions, possibly without direct access to the model, e.g., only a ``black-box'' access allowing certain queries. Explanations are typically generated post hoc and aim to answer questions such as why a particular decision was made, how it could have changed, or which inputs were most influential. Common forms include feature-attribution scores, counterfactual explanations, example-based explanations, and surrogate models. Explainability is particularly important when transparency by design is infeasible, but it also introduces new challenges, as explanations can themselves be misleading, unstable, or manipulable \mbox{\citep{Brughmans2024}}.

\item \textbf{Robustness} captures the ability of an ML system to maintain acceptable behavior under perturbations. These perturbations may occur in the training data or input, arising from noise, distributional shift, corrupted labels, or deliberate attacks. Techniques to enhance robustness range from detecting adversarial examples to training that accounts for the possibility of adversarial inputs or perturbed data. Robustness is especially critical in applications exposed to adversarial conditions or strict safety requirements, such as cybersecurity, aviation and air traffic control, autonomous driving, and other safety-critical systems.

\item \textbf{Fairness} concerns the prevention of unjustified disparities in model outcomes across individuals or groups defined by sensitive attributes, such as gender, race, or socioeconomic status. Bias can arise from historical data reflecting societal inequalities, as well as from data collection, labeling, or preprocessing practices. Addressing fairness may involve preprocessing strategies (e.g., reweighting or data augmentation), in-processing approaches that incorporate fairness criteria during training, or post hoc evaluation and verification of model behavior. Fairness notions are inherently multifaceted and sometimes mutually incompatible, requiring careful consideration of trade-offs among different definitions and their impact on predictive performance. Fairness is essential in domains such as hiring, lending, healthcare, and criminal justice, where biased decisions have significant societal consequences.

\item \textbf{Privacy} addresses the protection of sensitive information used or revealed by ML systems, including training data, individual records, and proprietary model information. Privacy risks may arise during training, inference, or through released artifacts such as model parameters, predictions, or explanations. Regulatory frameworks such as the General Data Protection Regulation (GDPR; \citealt{EU2016GDPR}) have elevated privacy to a legal requirement, pushing the development of techniques such as differential privacy, federated learning, and data minimization. Privacy is particularly critical in domains like healthcare, finance, and public administration, where ML systems routinely process personal or sensitive data.
\end{itemize}

\paragraph{Combinatorial Optimization (CO).}
This field focuses on decision-making problems defined over discrete, structured, or combinatorial solution spaces. In contrast to ML's use of the term ``model'', a CO \emph{model} is a precise mathematical formulation of a problem that specifies decision variables, objectives, and constraints. This formulation-centric viewpoint is central to the contributions surveyed in this paper.

A wide range of formal frameworks exists to express CO problems, including mixed-integer linear programming (MILP), semidefinite programming (SDP), constraint programming (CP), Boolean satisfiability (SAT), maximum satisfiability (MaxSAT), and satisfiability modulo theories (SMT). Many optimization problems arising in trustworthy ML (e.g., exact training under sparsity constraints, model pruning, robustness verification, fairness auditing, explanation generation, or privacy reconstruction) are NP-hard in general. This computational hardness precludes the existence of polynomial-time algorithms, unless P=NP, but it does not necessarily render these problems intractable in practice.

Indeed, modern CO solvers routinely handle large and complex instances by combining a wide range of algorithmic ingredients, including branching, bounding, cutting planes, presolving, decomposition methods, and internal heuristics. Crucially, progress in this area has been cumulative: no single technique explains the gains observed today, but decades of incremental improvements across these components have collectively pushed solution barriers. Experimental analyses show that, between the early 2000s and today, state-of-the-art MILP solvers have achieved speedups ranging from three to six orders of magnitude, with gains attributable to both algorithmic advances and hardware evolution \citep{Koch2022,Clautiaux2025}, and progress is still ongoing with the increased effective use of GPUs in linear-program solutions \citep{Applegate2021}. Problems that were entirely out of reach only a few decades ago are now solved to proven optimality or to tight optimality gaps within seconds or minutes. Even when an exact solution is infeasible, heuristic or approximate methods can be employed; notably, many standard ML algorithms themselves (such as gradient-based training, greedy tree induction, or local explanation methods) are heuristics. Adopting an optimization mindset makes this explicit and allows heuristic solutions to be assessed relative to known optima or bounds, at least for smaller instances. Moreover, for specific subclasses of problems, global optimality can often be achieved at scales sufficient for practical deployment. Pushing the boundary of what is solvable (and understanding where that boundary lies) has long been a defining objective of CO research, and it is precisely this perspective that motivates us to reexamine its growing role in trustworthy ML.

\section{CO for ML Transparency}
\label{sec:Transparency}

Transparency in ML is a broad desideratum that encompasses multiple, sometimes overlapping, interpretations. A widely used and practical proxy for transparency is \emph{model simplicity}: models that are compact, use fewer features, and involve fewer interactions are generally easier to inspect, understand, and audit. While classical training algorithms and feature selection methods can be steered towards simpler models, transparency-oriented objectives require a tighter integration of feature selection and model training, or the explicit inclusion of complexity measures (such as the $\ell_0$ norm, which counts the number of active features) directly in the learning objective. These formulations give rise to non-convex and NP-hard optimization problems. In this section, we review a range of CO-based approaches that address transparency from complementary angles. Some methods aim to learn sparse and interpretable models from scratch, others focus on simplifying or sparsifying already-trained models, and others explore the Rashomon set of near-optimal models to identify simpler alternatives without sacrificing predictive performance.

\subsection{Joint Learning and Feature Selection}
\label{subsec:Sparse}

Our analysis begins with learning methods that explicitly control model complexity by selecting a limited number of features.
For an initial dataset with $p$ features, there are $2^p$ possible feature subsets, making feature selection inherently combinatorial. Minimizing the number of selected features aligns naturally with Occam's razor, which suggests that simpler, more parsimonious models should be preferred whenever possible. Beyond improved interpretability, feature selection can enhance generalization, reduce computational costs, and limit data collection and storage requirements, an important consideration when features are expensive to acquire, as in biological experiments or medical diagnostics. Moreover, in line with the GDPR, data use should be ``adequate, relevant, and limited to what is necessary in relation to the purposes for which they are processed''. Accordingly, recent work on data minimization in ML has explicitly framed feature selection as a core mechanism for reducing unnecessary exposure of personal data \citep{Staab2025}.

Feature selection has been studied extensively for several decades; we refer readers to the survey of \citet{Guyon2003} for early works and classifications. Feature selection might be done (i) as a preprocessing step, (ii) iteratively while retraining, or (iii) completely integrated with the training of the model. We focus on the latter category, which involves optimization problems with explicit cardinality constraints or $\ell_0$ penalties, and has been a key focus in MILP research \citep{Tillmann2024}. Feature selection can improve model performance, prediction efficiency, and deliver a more interpretable model with additional insights into its features. In most cases, however, accuracy (measured through cross-validation or on a validation set) remains the dominant criterion \citep{Huang2015}.

Branch-and-bound (B\&B) techniques were introduced early in the feature selection literature to avoid exhaustive enumeration of all feature subsets \citep{Narendra1977}. These methods are known to be exact when the performance measure satisfies a monotonicity property with respect to the feature set, but also perform relatively well even when this assumption does not strictly hold \citep{Hamamoto1990}. In early approaches, the predictive model is treated largely as a black box, while most subsequent work leverages explicit mathematical formulations to design specialized learning algorithms. Without being exhaustive, we will discuss the development of such methods for two canonical cases: sparse linear regression and sparse support vector machines (SVMs).

\paragraph{Sparse linear regression.}
The best subset selection problem for linear regression, shown in Highlight~\ref{highlight:Regression}, is NP-hard and has become a canonical benchmark for joint learning and feature selection. Renewed interest in exact approaches was sparked by \citet{Bertsimas2016}, who formulated the problem as a mixed-integer quadratic program (MIQP) and demonstrated that, with modern solvers, even a compact big-$M$ formulation produced globally optimal sparse models on problems with hundreds to a few thousand features, yielding significant gains in accuracy and interpretability over convex surrogates such as the Lasso.

\begin{figure*}[!htbp] \refstepcounter{highlight}
\centering \scalebox{0.95}{
\begin{highlight}[ycornercontinuous]{Learning Sparse Linear Regression Models}
\label{highlight:Regression}
\begin{subequations}
\begin{align}
    \min_{\mathbf{w}} \quad 
    & \|\mathbf{y} - X\mathbf{w}\|_2^2 
      +  \lambda \,\|\mathbf{w}\|_2^2 \label{eq:ss_obj} \\
    \text{s.t.} \quad 
    & \|\mathbf{w}\|_0 \le \gamma \label{eq:ss_card}
\end{align}
\end{subequations}
Objective~\eqref{eq:ss_obj} combines the squared prediction error with an $\ell_2$-regularization term that improves numerical stability and controls the magnitude of the coefficients, while Constraint~\eqref{eq:ss_card} enforces a cardinality limit, restricting the model to at most $\gamma$ active features. Note that in several works, this constraint is replaced by an $\ell_0$ objective regularization term.
\end{highlight}}
\end{figure*}

Subsequent work focused on strengthening formulations and extending scalability through cutting planes and alternative reformulations. \citet{Bertsimas2020d} adopted a dual perspective that eliminates the regression coefficients and reformulates sparse regression as a pure binary convex optimization problem in the kernel space. This reformulation is solved via an outer-approximation algorithm that iteratively refines a MILP approximation using effective separation routines. In parallel, a line of work based on \emph{perspective reformulations} exploited the structure of the squared $\ell_2$ term to derive substantially tighter continuous relaxations for $\ell_0$-regularized regression problems \citep{Dong2015,Atamturk2020,Xie2020}. These formulations are typically expressed as mixed-integer second-order cone programs (MISOCPs) and yield tighter bounds. Additional computational gains have been achieved by designing custom B\&B schemes, as in \citet{Hazimeh2022,Guyard2024} and \citet{Liu2025a}. Finally, \citet{Meng2026} designed a B\&B algorithm that relies on perspective reformulation and the alternating direction method of multipliers (ADMM) to compute bounds, specifically tailored for GPU acceleration. This method achieves substantial speedups over CPU-based implementations on problems with thousands of samples and up to $p = 10^5$ features.

\paragraph{Sparse SVMs.}
Extensive work has also been conducted on sparse SVM formulations, with methodological developments closely paralleling those for sparse linear regression. SVMs with $\ell1$-norm regularization already promote sparsity intrinsically, so they were sometimes used directly as a feature selection algorithm \citep{Bi2003}. This, however, provides only weak, indirect control over the number of selected features. With the hinge loss, the resulting problem remains a linear program (LP) and is polynomially solvable, but large-scale instances may already require tailored algorithms based on column and constraint generation to be solved efficiently \citep{Dedieu2022}.

To obtain explicit guarantees on the number or cost of selected features, a series of works have formulated sparse SVM training as mixed-integer optimization problems with $\ell_0$-norm regularization or cardinality constraints. An early work by \citet{Chan2007} explored quadratic and semidefinite relaxation of this problem within a B\&B framework. Subsequently, \citet{Labbe2019} proposed MILP formulations for
feature selection in SVMs, relying on big-$M$ constraints and bound-tightening strategies. More recently, \citet{Bertsimas2021a} proposed an exact outer-approximation algorithm for $\ell_0$-regularized SVMs and related sparse classification problems, while \citet{Bomze2025} developed scalable conic decomposition methods for hard cardinality constraints, yielding tighter relaxations than classical big-$M$ models. Finally, \cite{Benitez-Pena2019a} and \cite{Guyard2024} also considered feature-specific acquisition costs, while \cite{Lee2022} developed a branch-and-cut-price methodology for group-wise feature acquisition.

\begin{figure*}[!htbp] \refstepcounter{highlight}
\centering \scalebox{0.95}{ \begin{highlight}[ycornerbinary]{Learning Sparse Support Vector Machines}
\label{highlight:SVM}
\begin{subequations}
\begin{align}
    \min_{\mathbf{w}, w_0, \boldsymbol{\xi}} \quad & \frac{1}{2} \|\mathbf{w}\|_2^2 + C \sum_{i=1}^{n} \xi_i \label{eq:svm_obj} \\
    \text{s.t.} \quad & y_i (\mathbf{w}^\top \mathbf{x}_i + w_0) \geq 1 - \xi_i, \qquad  & \forall i \in [n] \label{eq:svm_cons1} \\
   &\|\mathbf{w}\|_0 \leq \gamma \label{eq:svm_cons3} 
\end{align}
\end{subequations}
Objective~\eqref{eq:svm_obj} minimizes the sum of the  regularization term $\frac{1}{2} \|\mathbf{w}\|_2^2$ and hinge loss $C \sum_{i=1}^{n} \xi_i$, while Constraints~\eqref{eq:svm_cons1} ensure that classification errors are penalized via non-negative slack variables $\xi_i$.  Constraint~\eqref{eq:svm_cons3} limits the number of active features. 
\end{highlight}} \end{figure*}

\subsection{Dimensionality Reduction}

Dataset dimensionality reduction can also be achieved a priori through sparse principal component analysis (PCA). This technique is especially useful for high-dimensional data analysis, notably in bioinformatics, where individual raw features can be numerous, correlated, and difficult to interpret in isolation. Unlike feature selection, however, PCA constructs new derived features; therefore, sparsity aims to preserve some degree of transparency by ensuring that each component depends on only a limited subset of the original variables.

\begin{figure*}[!htbp] \refstepcounter{highlight}
\centering \scalebox{0.95}{ \begin{highlight}{Sparse Principal Component Analysis}
\label{highlight:PCA}
\begin{subequations}
\begin{align}
    \max_{\mathbf{w}} \quad & \mathbf{w}^\top A \mathbf{w}  \label{eq:pca_obj} \\
    \text{s.t.} \quad & \|\mathbf{w}\|_2 = 1, \quad \|\mathbf{w}\|_0 \leq \gamma \label{eq:pca_cons}
\end{align}
\end{subequations}

Objective~\eqref{eq:pca_obj} seeks to maximize the variance of the data explained by the (first) sparse principal component $\mathbf{w}$. In this formulation, $A = \frac{1}{n} X^\top X$ is the sample covariance matrix (assuming that each feature is centered to have zero mean). Constraints~\eqref{eq:pca_cons} enforce that the principal component $\mathbf{w}$ is normalized and sparse.
\end{highlight}} \end{figure*}

The literature on sparse PCA is vast and cannot be comprehensively reviewed here. Nevertheless, interest in the exact solution of this problem, or variants thereof, has been revived by recent improvements in mathematical optimization techniques. For the classical single-component formulation, \citet{Berk2019} developed a tailored B\&B algorithm based on linear algebraic bounds, while \citet{Dey2022} used $\ell_1$-relaxation and integer programming to obtain scalable dual bounds. In a complementary direction, \citet{Bertsimas2022a} and \citet{Li2024a} reformulated sparse PCA as mixed-integer semidefinite programs, deriving valid inequalities and cutting-plane or branch-and-cut algorithms, together with relax-and-round or approximation schemes for larger instances. More recently, \citet{Behdin2026} proposed a statistically motivated mixed-integer estimator based on the spiked covariance model, while \citet{CoryWright2026} extended the optimization perspective to sparse PCA with multiple orthogonal components, jointly optimizing several mutually orthogonal sparse principal components rather than selecting them one at a time.

\subsection{Training Interpretable Models}
\label{sec:train-interp}

There is no universal consensus on what makes a model interpretable, but it is generally agreed that rule-based methods, score-based models, regression models, and decision trees are comprehensible when their parameter count is limited. Following Occam's Razor, transparent model design seeks methods that produce inherently interpretable models with few parameters.

\paragraph{Scoring systems.}
Some of the easiest models to comprehend are scoring systems and, more broadly, generalized additive models (GAMs). These models deliberately avoid feature interactions: each feature contributes an additive score (via thresholds and small integer weights in scoring systems, or via learned univariate functions in GAMs), and the final prediction is obtained by summing these contributions and comparing to a threshold. Scoring models are especially common in domains such as credit scoring and healthcare, where they enable reliable decision-making with minimal computational support (e.g., the APACHE III ICU mortality score by \citealt{Knaus1991}).

Learning scoring systems is inherently combinatorial, as one seeks discrete and interpretable coefficients, sparsity, and strong predictive performance simultaneously.
The SLIM framework (Supersparse Linear Integer Models -- Highlight~\ref{highlight:SLIM}) proposed by \citet{Ustun2016} formalizes this task as an integer program that directly optimizes the $0$-$1$ loss and the $\ell_0$-norm of the coefficient vector, subject to integrality constraints. This approach yields sparse scoring systems with small coprime integer coefficients. Later work extended this framework to optimized risk scores by replacing the $0$-$1$ loss with logistic loss, enabling probabilistic predictions at the cost of solving a convex mixed-integer nonlinear program (MINLP) \citep{Ustun2019a}. To improve scalability and rapidly generate high-quality integer risk scores, \citet{Liu2022a} relied on beam search with tailored rounding heuristics while exploring a diverse set of near-optimal solutions. Finally, \citet{Molero-Rio2025} addressed scoring systems with continuous features by jointly selecting discretization thresholds and integer coefficients within a MINLP framework, thereby integrating feature binning and learning within a single formulation.

\begin{figure*}[!htbp] \refstepcounter{highlight}
\centering \scalebox{0.95}{ \begin{highlight}[ycornerbinary]{Learning Sparse Scoring Systems~\citep{Ustun2016}}
\label{highlight:SLIM}
\begin{subequations}
\begin{align}
    \min_{\mathbf{w}} \quad & \frac{1}{n} \sum_{i=1}^{n} \mathbb{1}[ y_i \mathbf{w}^\top \mathbf{x}_i \leq 0 ] + \lambda\|\mathbf{w}\|_0 + \epsilon \|\mathbf{w}\|_1 \label{eq:slim_obj} \\
    \text{s.t.} \quad & \mathbf{w} \in \Psi^p  \label{eq:slim_cons} 
\end{align}
\end{subequations}

Objective~\eqref{eq:slim_obj} minimizes the 0-1 loss over all the training samples and penalizes the $\ell_0$-norm of the coefficients' vector $\mathbf{w}$ to encourage sparsity. It additionally considers the $\ell_1$-norm $\| \mathbf{w}\|_1$ with a small coefficient to break ties in favor of solutions with coprime values. Constraints~\eqref{eq:slim_cons} restrict the coefficients to belong to a predefined set of integer values $\Psi$.
\end{highlight}} \end{figure*}

\paragraph{Rule sets.} These models offer an alternative strategy for rule-based binary classification, where a data point is classified as positive if it satisfies at least one rule in the set, and negative otherwise. They are unordered and naturally represented in disjunctive normal form (DNF). Early approaches to learning rule sets typically relied on heuristic rule mining followed by greedy selection. Later methods formulated the problem as a CO task. Notably, \citet{Kamath1992} proposed a SAT-based formulation and a MILP model solved via an interior-point method for constructing minimal disjunctive rule sets consistent with the training data. More recently, \citet{Dash2018a} and \citet{Lawless2023a} proposed MIP approaches that explicitly optimize both predictive accuracy and model sparsity, minimizing a Hamming loss proxy for $0$-$1$ classification error while constraining model complexity (Highlight~\ref{highlight:Boolean-Rule}). Due to the exponential number of candidate rules, these methods leverage column generation, iteratively expanding the rule set by solving a pricing problem to identify new candidates.

\begin{figure*}[!htbp] \refstepcounter{highlight}
\centering \scalebox{0.95}{
\begin{highlight}[ycornerbinary]{Learning Boolean Rule Sets \citep{Lawless2023a}}
\label{highlight:Boolean-Rule}
\begin{subequations}
\begin{align}
    \min_{\mathbf{u}, \boldsymbol{\xi}} \quad & \sum_{\substack{i=1\\ \text{s.t.}~y_i = 1}}^{n} \xi_i + \sum_{\substack{i=1\\ \text{s.t.}~y_i = -1}}^{n} \sum_{k \in \mathcal{K}_i} u_k \label{eq:bool_obj} \\
    \text{s.t.} \quad & \xi_i + \sum_{k \in \mathcal{K}_i} u_k \geq 1 & \forall i \in [n]~\text{s.t.}~y_i = 1\label{eq:bool_cons1} \\
    & \sum_{k \in \mathcal{K}} c_k u_k \leq \gamma \label{eq:bool_cons_complexity} 
\end{align}
\end{subequations}

This formulation minimizes the Hamming loss, a surrogate for $0$-$1$ classification error. Each binary variable $u_k$ indicates whether rule $k \in \mathcal{K}$ is selected in the rule set, and binary variables $\xi_i$ flag misclassified positive examples. For each negative example $i$, the model incurs a penalty equal to the number of selected rules in $\mathcal{K}_i$, where $\mathcal{K}_i$ denotes the set of rules that cover~$i$.
Constraint~\eqref{eq:bool_cons1} ensures that positive examples are covered by at least one rule or are misclassified otherwise, while \eqref{eq:bool_cons_complexity} enforces an upper bound on the total complexity of the selected rules ($c_k$ usually depending on the number of conditions involved in rule $k$). Typically, $\mathcal{K}$ is large and requires the use of column generation.
\end{highlight}}
\end{figure*}

\paragraph{Rule lists.} In contrast to scoring systems and rule sets, which do not set an order, rule-list models use an ordered succession of ``if then'' rules to classify data.
$$
\left|\begin{array}{l}
\textbf{IF } \text{Condition 1} \textbf{ THEN } \text{Label 1} \\
\textbf{ELSE IF } \text{Condition 2} \textbf{ THEN } \text{Label 2} \\
\vdots \\
\textbf{ELSE } \text{Default Label}
\end{array}\right.
$$
These models can also be trained using mathematical programming to optimize prediction accuracy and sparsity. 
\citet{Rudin2018} introduced a general MILP framework that formalizes this task as two subsequent models: one for mining candidate rules from data using support and confidence measures, and another for selecting and ordering a subset of these rules under sparsity and accuracy constraints. In contrast, \citet{Angelino2018} proposed CORELS, a specialized B\&B algorithm designed to efficiently find the globally optimal ordering of a fixed, pre-mined set of binary rules. The method relies on bounding and symmetry-elimination techniques. While it scales well to datasets with hundreds of thousands of examples, its applicability remains limited to low-dimensional settings, as the number of potential decision rules grows exponentially with the number of features.

Finally, \citet{Yu2021} developed SAT and MaxSAT formulations for learning rule sets and lists that are optimal with respect to the total number of literals, rather than the number of rules (see Highlight~\ref{highlight:MaxSAT}). Their framework supports both \emph{perfect classifiers}, which aim for zero training error, and \emph{sparse variants} that balance accuracy and complexity via regularization. Unlike earlier methods that rely on rule mining as a preprocessing step, it constructs rules dynamically during the solution process.
More recently, \citet{Rober2025} introduced a large-scale rule-generation method built on column generation. Essentially, the method builds an additive ensemble in which each base learner is a rule. Their approach formulates rule learning as an LP whose columns correspond to candidate rules, and iteratively solves a pricing subproblem to generate new rules with negative reduced cost. The exact pricing problem is shown to be NP-hard, and the authors proposed a fast decision-tree-based heuristic as a proxy pricing method to ensure scalability. This framework supports additional linear constraints for interpretability (e.g., rule-length penalties) and fairness (e.g., disparate mistreatment metrics), demonstrating good performance across large datasets.

\begin{figure*}[!htbp] \refstepcounter{highlight}
\centering \scalebox{0.95}{
\begin{highlight}[ycornermulticlass]{Learning Sparse Decision Lists with MaxSAT~\citep{Yu2021}}
\label{highlight:MaxSAT}
\begin{subequations}
\begin{align}
    &\min_{\mathbf{u}, \mathbf{s}, \boldsymbol{\eta}, \mathbf{v}, \boldsymbol{\xi}} \quad  \sum_{i=1}^{n} \xi_i + \lambda \sum_{j=1}^{m} (1 - u_j)  \label{eq:maxsat_obj} \\
    & u_j + \sum_{r \in (\mathcal{K} \cup \mathcal{Y})} s_{jr} = 1 & \forall j \in [m] \label{eq:maxsat_node_use} \\
    & u_j \Rightarrow u_{j+1} & \forall j \in [m-1] \label{eq:maxsat_order} \\
    & \eta_{i1} = v_{i1} = 1 & \forall i \in [n] \label{eq:unclassified_init} \\
    & \eta_{i,j+1} \Leftrightarrow \eta_{ij} \land \left[ \left( \bigwedge_{y \in \mathcal{Y}} \neg  s_{jy} \right)   \lor  \neg v_{ij} \right] & \forall i \in [n], j \in [m-1] \label{eq:not_classified_update} \\
    & v_{i,j+1} \Leftrightarrow \left( \bigvee_{y \in \mathcal{Y}} s_{jy} \land \eta_{i,j+1} \right) \lor \left( v_{ij} \land \bigvee_{r \in \mathcal{K}} (s_{jr} \land a_{ir}) \right) & \forall i \in [n],\; j \in [m-1] \label{eq:validity_update} \\
    & \left( s_{jy} \land v_{ij} \right) \Rightarrow \xi_i & \forall i \in [n],\; j \in [m],\; y \in \mathcal{Y} \setminus \{y_i\} \label{eq:correct_prediction_or_misclassify} \\
    & \bigvee_{j \in [m]} \left( \bigvee_{y \in \mathcal{Y}} s_{jy} \land v_{ij} \right) \lor \xi_i & \forall i \in [n] \label{eq:must_be_classified_or_misclassified}\\
    & u_{j+1} \Rightarrow u_j \lor \bigvee_{y \in \mathcal{Y}} s_{jy} & \forall j \in [m-1] \label{eq:last_node_is_leaf_1} \\
    & u_m \lor \bigvee_{y \in \mathcal{Y}} s_{my} \label{eq:last_node_is_leaf_2}
\end{align}
\end{subequations}

This formulation constructs sparse decision lists by minimizing the number of misclassified training examples ($\sum \xi_i$) and the total number of used literals, penalized by a regularization parameter $\lambda$. These objectives are encoded as soft clauses within a weighted partial MaxSAT problem. The model is defined over a fixed number $m$ of decision nodes and a set $\mathcal{K}$ of binary conditions on features' values. Each node~$j$ is either active ($u_j = 0$) and selects exactly one literal $r$ (either a feature condition or a class assignment) by setting $s_{jr}$ to \textsc{True}, or it is inactive ($u_j = 1$), as enforced by~\eqref{eq:maxsat_node_use}. Contiguity is guaranteed by~\eqref{eq:maxsat_order}, ensuring that no unused node precedes a used one.
Constraint~\eqref{eq:unclassified_init} initializes every example $i$ as unclassified ($\eta$ variables) and valid ($v$ variables) at the first node. 
The update constraints~\eqref{eq:not_classified_update} and~\eqref{eq:validity_update} define whether each example $i$ remains unclassified or continues to be valid at subsequent nodes.
Constraint~\eqref{eq:not_classified_update} states that $i$ remains unclassified at node $j+1$ iff it was previously unclassified and either $j$ is not a class node, or $i$ was invalid at the previous node (so $i$ is not classified by the rule that ends at $j$).
Constraint~\eqref{eq:validity_update} enforces that $i$ is valid at node $j+1$ iff node $j$ was a class node and $i$ remains unclassified after it (therefore resets validity for the next rule) or if $i$ was valid at node $j$ and satisfies the condition selected at this node (therefore $i$ remains valid for the rule being built). 
Here, $a_{ir}$ is a boolean constant indicating whether example $i$ satisfies condition $r$. 
Constraint~\eqref{eq:correct_prediction_or_misclassify} enforces that if an example is classified by a rule, then it must be correctly classified (i.e., predicted class must equal $y_i$) or incur a misclassification penalty. Constraint~\eqref{eq:must_be_classified_or_misclassified} ensures that every example is either covered by some rule or marked as misclassified.
Finally, Constraints~\eqref{eq:last_node_is_leaf_1} and~\eqref{eq:last_node_is_leaf_2} state that the last used node must be a class assignment.
Note that this model can be simplified to learn sparse decision sets by removing the $\eta_{ij}$ variables (which track which examples were classified by previous rules) and the constraints that compute them. 
\end{highlight}}
\end{figure*}

\paragraph{Decision trees.}
Among all sparse and interpretable models trainable via CO, optimal decision trees (ODTs) have received the most sustained research attention~\citep{Costa2023}. Decision trees are valued for their transparency, as their interpretability is directly tied to structural parameters such as depth and the number of leaves. While greedy heuristics like CART have long been the standard, their myopic decisions often lead to suboptimal trees, both in terms of generalization accuracy and simplicity. Several studies reported improvements of a few percentage points in out-of-sample accuracy when switching from greedy to globally optimal constructions \citep{Bertsimas2017, Demirovic2022, VanderLinden2024}. Beyond accuracy, global optimization also supports additional constraints and objectives, notably cost-sensitive variants \citep[see, e.g.][]{Lomax2013} in which individual splits (tests) incur heterogeneous costs, as is common in medical diagnosis and sequential decision-making settings.

Broadly, two methodological families can be distinguished. The first, most versatile, relies on mathematical programming and includes models by \citet{Bertsimas2017}, \citet{Verwer2019}, \citet{Gunluk2021}, \citet{Ales2024}, and \citet{Aghaei2024}. Notably, the flow-based formulation by \citet{Aghaei2024} (see Highlight~\ref{highlight:ODT}) avoids big-$M$ constraints to achieve a tighter linear relaxation and can be exploited within a Benders decomposition approach. \citet{Ales2024a} further improved scalability through iterative sample aggregation and disaggregation strategies, and \citet{Firat2020a} and \citet{Patel2024} explored column-generation approaches to scale to a larger number of rows (i.e., data points). Alternative formulations based on CP \citep{Verhaeghe2020a}, SAT \citep{Narodytska2018,Shati2023,Schidler2024a} and MaxSAT \citep{Hu2020} have also been proposed.

\begin{figure*}[!htbp] \refstepcounter{highlight}
\centering \scalebox{0.95}{ 
\begin{highlight}[ycornermulticlass]{Learning Optimal Decision Trees~\citep{Aghaei2024}}
\label{highlight:ODT}
\begin{subequations}
\begin{align}
    \max_{\mathbf{z}, \mathbf{b}, \mathbf{w}} \quad & \sum_{i \in [n]} \sum_{j \in \mathcal{L}} z^i_{j, t} \label{eq:odt_obj} \\
    \text{s.t.} \quad 
    & \sum_{f \in [p]} b_{j,f} = 1 & \forall j \in \mathcal{B} \label{eq:odt_branching} \\
    & \sum_{y \in \mathcal{Y}} w_{j,y} = 1 & \forall j \in \mathcal{L} \label{eq:odt_predict_one} \\
    & z^i_{s, 1} \leq 1 & \forall i \in [n] \label{eq:odt_source} \\
    & z^i_{a(j), j} = z^i_{j, \ell(j)} + z^i_{j, r(j)} & \forall j \in \mathcal{B},\; \forall i \in [n] \label{eq:odt_flow1} \\
    & z^i_{a(j), j} = z^i_{j, t} & \forall j \in \mathcal{L},\; \forall i \in [n] \label{eq:odt_flow2} \\
    & z^i_{j, \ell(j)} \leq \sum_{f \in [p] : x_{if} = 0} b_{j,f} & \forall j \in \mathcal{B},\; \forall i \in [n] \label{eq:odt_left} \\
    & z^i_{j, r(j)} \leq \sum_{f \in [p] : x_{if} = 1} b_{j,f} & \forall j \in \mathcal{B},\; \forall i \in [n] \label{eq:odt_right} \\
    & z^i_{j, t} \leq w_{j, y_i} & \forall j \in \mathcal{L},\; \forall i \in [n] \label{eq:odt_correct}
\end{align}
\end{subequations}

This formulation constructs an ODT of fixed depth for binary datasets ($\mathcal{X} = \{0,1\}^p$) by maximizing the number of correctly classified examples.
The tree is modeled as a directed acyclic graph between a source $s$ and a sink $t$. For each branching node $j \in \mathcal{B}$, binary variable $b_{j,f}$ indicates whether it branches on feature $f \in [p]$, and Constraint~\eqref{eq:odt_branching} ensures that $f$ is unique. For each leaf $j \in \mathcal{L}$, binary variable $w_{j,y}$ indicates whether it predicts class $y$, and Constraint~\eqref{eq:odt_predict_one} ensures that $y$ is unique. 
Finally, for each $j \in \mathcal{B} \cup \mathcal{L}$, binary flow variable $z^i_{a(j),j}$ indicates whether example $i$ is routed from $a(j)$ to $j$.
Flow initialization and conservation are enforced through~\eqref{eq:odt_source}-\eqref{eq:odt_flow2}, where $a(j)$, $\ell(j)$, and $r(j)$ are respectively the ancestor, left child, and right child of $j$. 
Constraints~\eqref{eq:odt_left} and~\eqref{eq:odt_right} determine whether an example proceeds to the left or right child based on the feature selected at node $j$ and the value of $\smash{x_{if}}$. Constraint~\eqref{eq:odt_correct} ensures that example $i$ can only flow to a leaf if it predicts its true label $y_i$. Therefore, only correctly classified examples have non-zero flow in~\eqref{eq:odt_obj}.
\end{highlight}} 
\end{figure*}

The second family of methods combines dynamic programming (DP) with B\&B, abbreviated as DPB\&B. DP-based methods typically exploit caching of subproblem solutions, so their computational effort is primarily driven by feature-space dimensionality and the number of possible feature levels. Early contributions include DL8.5 \citep{Aglin2020}, OSDT \citep{Hu2019a}, GOSDT \citep{Lin2020a}, and MurTree \citep{Demirovic2022}, all of which rely on tailored bounding procedures and caching strategies. More recently, \citet{Demirovic2023} introduced BLOSSOM, an anytime DPB\&B algorithm that leverages efficient bounds, heuristic feature orderings, and caching strategies. \citet{vanderLinden2023} further generalized this line of work with STreeD, a DP framework for optimal decision tree learning with separable objectives and constraints. They provide necessary and sufficient conditions under which such objectives and constraints can be optimized by DP, thereby extending previous frameworks beyond standard additive objectives.
To improve memory use and runtime, \citet{Chaouki2025} formulated the search for optimal sparse decision trees as an AND-OR graph search and employed a specialized AO*-type search strategy.  Complementarily, \citet{Brita2025} proposed a specialized method for numerical features with many possible split values. Hybrid strategies are also gaining traction. Notably, \citet{Schidler2024a} proposed a two-stage approach in which heuristically trained trees are locally refined via SAT-based optimization. \citet{Babbar2025} presented SPLIT, a lookahead framework that applies exact optimization to the upper levels of the tree and reverts to greedy splitting near the leaves. This strategy yields near-optimal trees while achieving substantial speed-ups.

Beyond traditional trees, optimal decision diagrams (ODDs) generalize the ODT formalism by allowing flow merging, whereby branches and sample paths can join as they descend through the diagram. This enables ODDs to compactly express more complex relationships (e.g., XOR patterns or cardinality constraints) that would require an exponential number of leaves in standard decision trees. Moreover, by enabling flow merging, these models mitigate the sample sparsity problem that arises when deeper trees produce many underpopulated leaf paths. ODDs can be trained using MaxSAT-based approaches \citep{Hu2022}, MILP formulations \citep{Florio2023a}, or B\&B \citep{Komusiewicz2025}. 

Finally, it is worth noting that stand-alone decision trees (whether learned heuristically or via global optimization) are rarely competitive for complex prediction tasks. In such settings, ensemble methods such as random forests and gradient-boosted trees remain the dominant approach for strong predictive performance. A natural question is whether replacing heuristic base learners with optimal ones can further improve ensemble accuracy. The answer is nuanced, based on empirical evidence: as shown by \citet{Akkerman2025}, ensembles built from ODTs do not consistently outperform those built from greedy learners such as CART, possibly due to reduced diversity.

\paragraph{Binarized neural networks (BNNs).} In those networks, both weights and activations are restricted to binary values (typically $\{-1, +1\}$ or $\{0, 1\}$), which can offer advantages in memory efficiency, inference speed, and hardware compatibility. However, their training remains challenging due to the non-differentiability of binary operations, which hinders the use of standard gradient-based methods. Early work by \citet{Hubara2016} addressed this using the straight-through estimator (STE), a heuristic that permits backpropagation through binary units but often yields suboptimal solutions. Some CO methods have also been explored. \citet{ToroIcarte2019} proposed hybrid MIP and CP formulations that learn BNNs while optimizing proxies for generalization, including network sparsity and margin robustness. More recently, \citet{Aspman2024} introduced a formulation grounded on nonsmooth analysis, casting BNN training as the solution to a subadditive dual of a conic MIP. Their approach leverages the theory of tame functions, allowing differentiation through discrete layers using tools from variational analysis. Despite these advances, such approaches remain limited to networks far smaller than those that can be trained with conventional deep learning pipelines.

\paragraph{Symbolic regression.} This task seeks to identify concise mathematical expressions that explain observed data without assuming a fixed functional form \citep{Dong2025}. Classical approaches rely on genetic programming to evolve expressions as trees \citep{Koza1992}. Recent advances also include learning-based techniques and mathematical programming formulations. Notably, \citet{Cozad2018} and \citet{Austel2020} proposed global MINLP frameworks that address symmetries and redundancies in expression search. Those methods were later extended by \citet{Engle2022} to rely on derivative information. Further refinements by \citet{Neumann2020} and \citet{Kim2023} reduced the number of binary variables, leading to stronger formulations.


\subsection{Model Compression}

Whereas the previous section focused on learning an interpretable model from scratch, in many situations, an initial model is available. When this happens, it is often advantageous to reuse the model and attempt to \emph{simplify} it without altering its functionality.

\paragraph{Neural network pruning.} Pruning has been the subject of extensive research due to its central role in reducing memory and energy usage, accelerating inference \citep{Cheng2024}, and facilitating integration in optimization and decision pipelines \citep{Huchette2023}. Pruning strategies for neural networks are typically classified as either \emph{unstructured}, which removes individual weights, or \emph{structured}, which eliminates entire computational blocks such as neurons, channels, or attention heads. Due to the vectorized nature of GPU hardware, unstructured sparsity often fails to yield practical speedups without specialized compilers or hardware, whereas structured pruning yields more consistent performance gains.

While the scale of modern deep networks limits the direct application of CO, several CO-assisted pruning methods have met empirical success. Early work by \citet{Serra2020,Serra2021} introduced MILP formulations to identify \emph{stable} neurons in rectified linear unit (ReLU) networks, allowing exact simplification without altering their prediction. In parallel, \citet{Zhang2018b} and \citet{Carreira-Perpinan2018a} tackled the $\ell_0$-norm pruning problem through Lagrangian reformulations, using the alternating direction method of multipliers. \citet{Cacciola2023} applied a perspective reformulation to each group of weights (e.g., filters or channels), yielding a tighter continuous relaxation and leading to the derivation of a novel regularization term referred to as the structured perspective regularizer. It generalizes the Lasso to the structured setting, encourages group-wise sparsity, and integrates seamlessly into standard training procedures via gradient-based optimization.

Taking another direction, \citet{Yu2022} revisited the classical Optimal Brain Surgeon (OBS) framework of \citet{Hassibi1993a}, a pruning method that uses a local quadratic approximation of the loss to assess the effect of pruning individual weights. As formalized in Highlight~\ref{highlight:CBS}, the goal is to minimize the second-order Taylor approximation of the loss under a sparsity constraint. Unlike diagonal approximations used in OBS, this variant retains off-diagonal Hessian terms, capturing weight interactions. However, solving the resulting quadratic integer programming problem is challenging. To improve scalability, \citet{Benbaki2023} proposed CHITA, a framework that replaces the dense Hessian with a lower-rank approximation, reducing the problem to a cardinality-constrained least-squares regression task. Then, \citet{Meng2024} incorporated FLOP-aware constraints to better reflect deployment costs, and \citet{Meng2024a} developed a general framework for structured pruning that targets entire neurons, channels, or attention heads. This method achieves state-of-the-art pruning performance on large-scale vision and language models, scaling to tens of billions of parameters through local search over quadratic integer formulations.

\begin{figure*}[!htbp] \refstepcounter{highlight}
\centering \scalebox{0.95}{
\begin{highlight}{Combinatorial Brain Surgeon for Neural Network Pruning~\citep{Yu2022}}
\label{highlight:CBS}
\begin{subequations}
\begin{align}
    \min_{\mathbf{w}, \mathbf{u}} \quad & \frac{1}{2} (\mathbf{w} - \bar{\mathbf{w}})^\top \nabla^2 L(\bar{\mathbf{w}}) (\mathbf{w} - \bar{\mathbf{w}}) \label{eq:cbs_obj} \\
    \text{s.t.} \quad & u_j = 1 \Rightarrow w_j = 0  &  \forall j \in [m] \label{eq:cbs_pruned} \\
    & \sum_{j=1}^m u_j = \gamma\label{eq:cbs_cardinality} 
\end{align}
\end{subequations}

This formulation minimizes a second-order approximation of the loss function around the trained weights $\bar{\mathbf{w}}$, using the Hessian $\nabla^2 L(\bar{\mathbf{w}})$. For $j \in [m]$, binary variable $u_j$ indicates whether weight $w_j$ is pruned ($u_j = 1$) or not. Constraint~\eqref{eq:cbs_pruned} sets pruned weights to zero, while Constraint~\eqref{eq:cbs_cardinality} enforces a target sparsity level.
\end{highlight}}
\end{figure*}

\paragraph{Pruning ensemble models.}
Ensemble models (e.g., random forests or gradient boosting) remain widely used for tabular data due to their strong predictive performance. However, like neural networks, they suffer from slower inference and limited interpretability due to their larger size. This has motivated efforts to simplify ensemble models, including several approaches based on mathematical programming, to improve inference speed and interpretability. Early works include the MIQP formulation by \citet{Zhang2006}, which optimizes ensemble accuracy and diversity using quadratic optimization, later extended by \citet{Li2012a} and \citet{Cavalcanti2016} to consider different diversity measures. Later on, \citet{Liu2023} proposed a framework that simplifies regression tree ensembles by trimming depth layers of individual trees rather than removing entire trees, formulating the problem as a regularized least-squares optimization with additional constraints. Additionally, \citet{Devos2025} studied joint ensemble pruning, trimming, and reweighting.

Recently, \citet{Emine2025} and \citet{Carrizosa2025} proposed two related formulations. The former considers the task of \emph{faithful pruning}, where the goal is to reduce the size of a tree ensemble while performing additional reweighting, without altering its prediction function. This requirement, analogous to lossless pruning in neural networks \citep{Serra2020} or faithful model compression via born-again trees \citep{Vidal2020a}, is enforced by functional equivalence constraints of the form $g(\mathbf{x}) = h(\mathbf{x})$ for all inputs $\mathbf{x}$. Although this yields an optimization problem with exponentially many constraints, it can be solved efficiently using a cutting-plane algorithm that iteratively adds counterexamples where the pruned and original models disagree (see Highlight~\ref{highlight:FIPE}). In contrast, \citet{Carrizosa2025} studied ensemble pruning from an empirical risk perspective, aiming to balance sparsity and accuracy over a finite dataset while also incorporating fairness constraints. Their formulation can be interpreted as a sparse soft-margin SVM in a meta-feature space, where each feature corresponds to the prediction of a base learner. Subsequent studies built on the approach of \citet{Emine2025} to enhance pruning capabilities. \citet{Yajima2026} restricted the faithfulness guarantee to an in-distribution region, identified by some plausibility score calibrated to encompass a chosen fraction of the empirical data. In another work, \citet{Akkerman2026} restricted faithfulness not only to an in-distribution region but also to high-confidence regions of the original ensemble. Their method speeds up the cutting-plane algorithm by using a tailored CP formulation and leveraging column generation to iteratively add new trees to the ensemble, further enhancing compression capabilities (e.g., when fewer trees can replace multiple original trees in the region of interest).

\begin{figure*}[!htbp] \refstepcounter{highlight}
\centering
\scalebox{0.95}{
\begin{highlight}[ycornermulticlass]{Functionally Identical Pruning of Ensembles \citep{Emine2025}}
\label{highlight:FIPE}
\begin{subequations}
\begin{align}
    \min\limits_{\mathbf{w} \geq \mathbf{0}} \quad & \|\mathbf{w}\|_1 \label{eq:fipe_obj} \\
    \text{s.t.} \quad & \sum_{j=1}^{m} w_j \left( h^{H(\mathbf{x})}_j(\mathbf{x}) - h^y_j(\mathbf{x}) \right) \geq 1 
    & \forall \mathbf{x} \in \mathcal{X},\; \forall y \in \mathcal{Y} \setminus \{H(\mathbf{x})\} \label{eq:fipe_cons1}
\end{align}
\end{subequations}

Objective~\eqref{eq:fipe_obj} minimizes the $\ell_1$-norm of the weight vector $\mathbf{w}$, a convex surrogate that is empirically observed to encourage sparsity. For each tree $j \in [m]$, $w_j$ denotes its weight in the reweighted ensemble, with $h_j(\mathbf{x}) \in \mathcal{Y}$ its prediction and $h_j^y(\mathbf{x})$ its confidence score for class $y$.
Constraints~\eqref{eq:fipe_cons1} enforce \emph{functional equivalence}: for every input $\mathbf{x} \in \mathcal{X}$ and every label $y$ differing from the original ensemble prediction $H(\mathbf{x})$, the margin of the pruned model must still favor the correct class by at least 1. This ensures that the pruned model retains the original ensemble’s predictions across the entire input space. Since $\mathcal{X}$ represents a high-dimensional feature space, the model is solved via a cutting-plane algorithm that iteratively identifies counterexamples (inputs for which predictions diverge) and adds corresponding constraints until full equivalence is achieved.
\end{highlight}} \end{figure*}

\paragraph{Boosting from mathematical programming lenses.}
The connection between training and reweighting of boosting ensembles for classification and SVM models discussed in the previous paragraph has even deeper roots.
Although AdaBoost, a seminal boosting method, does not explicitly maximize margins, it has long been observed that it implicitly improves them by effectively minimizing an exponential loss through coordinate descent \citep[see, e.g.,][]{Friedman2000}.
This insight motivated the development of \emph{totally corrective boosting} methods, which recast ensemble learning as a column generation procedure: unlike AdaBoost, which adds one weak learner at a time without revisiting previous coefficients, the master problem jointly re-optimizes the weights of all selected weak learners, while the pricing problem identifies new base learners that improve classification margins or related objectives \citep{Grove1998,Demiriz2002,Shen2010}. The dual variables of the master problem naturally act as adaptive sample weights, emphasizing hard or misclassified examples.

A seminal methodology along this line, LPBoost \citep{Grove1998,Demiriz2002}, maximizes the minimum margin over the training set using a soft-margin SVM (see Highlight~\ref{highlight:LPBoost}). Unlike standard SVMs that operate on raw input features, LPBoost operates in the space of weak learners: each weak learner is a feature, and the final classifier is a linear combination of them. This method, however, might suffer from degeneracy because it focuses on the worst margin. To improve robustness, \citet{Shen2010} proposed MD-Boost, which regularizes the margin distribution by maximizing its mean and minimizing its variance via a quadratic program. Several follow-up approaches further improved generalization by regularizing the margin distribution, either through quadratic penalties, as in MD-Boost, or entropy-based smoothing as in Soft-Boost and ERLPBoost \citep{Ratsch2007,Warmuth2008b}. Most recently, \citet{Akkerman2025} introduced two new variants: NM-Boost, which explicitly penalizes negative margins to better handle misclassifications, and QRLP-Boost, which combines entropy and quadratic regularization for improved stability. Through extensive empirical investigation, they observed that totally corrective methods (such as NM-Boost) can match or exceed the performance of popular boosting approaches (such as XGBoost and LightGBM) when using shallow base learners, yielding sparser and more interpretable ensembles. However, these methods remain sensitive to hyperparameters and do not scale well when using deeper base learners.

\begin{figure*}[!htbp] \refstepcounter{highlight}
\centering
\scalebox{0.95}{
\begin{highlight}[ycornerbinary]{Totally Corrective Boosting~\citep{Demiriz2002}}
\label{highlight:LPBoost}
\begin{subequations}
\begin{align}
\max\limits_{\mathbf{w} \geq \mathbf{0},\rho} \quad & \rho \label{eq:lpboost_obj}\\
\text{s.t.} \quad & y_i \sum_{j=1}^{m} w_j h_j(\mathbf{x}_i) \geq \rho, \quad \forall i \in [n] \label{eq:hard_margin}\\
 &\|\mathbf{w}\|_1 = 1
\end{align}
\end{subequations}
Objective~\eqref{eq:lpboost_obj} maximizes the minimum margin $\rho$, which is computed in~\eqref{eq:hard_margin} based on the weights $\mathbf{w}$ associated with the $m$ current base learners. The dual values of this master problem represent weights for the training examples in a pricing problem, which is solved to potentially identify new base learners that are added to the master formulation.
\end{highlight}} \end{figure*}


\subsection{Exploration of Rashomon Sets}

Presenting a single model to a practitioner is often insufficient; many prediction tasks admit a multitude of models with equivalent or near-equivalent empirical accuracy, collectively known as the \emph{Rashomon set}~\citep{Breiman2001a}. Formally, this set consists of models whose empirical loss lies within a small tolerance $\epsilon$ of the minimum achievable loss. These models may be indistinguishable in predictive performance, yet they differ substantially in structure, sparsity, fairness, or even individual predictions. As highlighted in \citet{Marx2020}, such predictive multiplicity can lead to high disagreement rates among equally accurate classifiers, raising concerns about arbitrariness while also creating opportunities to select models with desirable properties. For this reason, \citet{Rudin2024} argued that model reporting should extend beyond a single fitted solution and reflect the landscape of ``many good models.''

CO provides principled approaches for exploring Rashomon sets. Enumeration-based methods leverage B\&B to explicitly list or sample near-optimal models within a hypothesis class, such as sparse decision trees \citep{Xin2022} or rule-set classifiers \citep{Ciaperoni2024}, enabling downstream analyses of stability, fairness, or feature attribution. While exhaustive enumeration enables complete characterization, its output quickly becomes intractable. To address this issue, \citet{Arslan2025} enumerated the Rashomon set of sparse decision trees in nondecreasing order of objective value, enabling early stopping once enough high-quality candidates have been generated.

Enumeration-free approaches instead characterize extremal properties of the Rashomon set without explicitly listing its elements. For instance, in generalized additive models, \citet{Chen2023a} optimized convex surrogates to bound feature importance under accuracy constraints, while \citet{Langlade2025} proposed MILP formulations that use the true 0-1 loss and explicit sparsity budgets to compute certifiable ranges for fairness (and other desiderata) across scoring systems and decision diagrams. Complementing these, \citet{Liu2025} introduced MOSS, a multi-objective optimization framework for rule-set models that maps the Pareto frontier across accuracy, sparsity, and stability.

Several heuristic or approximate methods have also been proposed to explore Rashomon sets at scale. Although they do not certify exhaustive coverage, they trade completeness for substantial gains in runtime and memory efficiency, while often recovering a large fraction of the exact set in practice. For decision trees,~\citet{Babbar2025} introduced a scalable approximation method that hybridizes greedy heuristics with B\&B search, while~\citet{Heile2025} proposed a lighter recursive scheme based on heuristic completions, substantially reducing computational cost while recovering nearly all of the exact Rashomon set empirically.
More recently,~\citet{Heile2026} introduced PRAXIS, a proxy-guided AND/OR-graph search that uses proxy completions to prune candidate splits and budget refinement to allocate the admissible loss budget across child subproblems. Experiments show that PRAXIS recovers a larger fraction of the Rashomon set than previous approximation methods, while remaining orders of magnitude faster and more memory-efficient than exact enumeration.

            \section{CO for ML Explainability}

Whereas the previous section focused on building or improving transparent models, we now assume a trained predictor and aim to \emph{explain} its decisions to end users. The distinction between \emph{interpretability by design} and \emph{post hoc explainability} is especially salient in high-stakes applications: as advocated in \citet{Rudin2019}, inherently interpretable models as covered in the previous section should be preferred whenever possible, but there are many situations where more complex models provide better predictive performance, and where post hoc explanation techniques are also needed.

In this section, we organize explanation techniques into four main families: \emph{feature-subset explanations}~(\S\ref{sec:exp:feature-subset}), which characterize predictions in terms of minimal subsets of input features and include counterfactual, contrastive, and abductive explanations; \emph{feature-attribution explanations}~(\S\ref{sec:exp:feature-attrib}), which assign quantitative responsibility scores to features, such as Shapley values or related attribution methods; \emph{prototype-based explanations}~(\S\ref{sec:exp:feature-proto}), which justify predictions by pointing to representative or influential instances in the training set; and finally, \emph{global surrogate explanations}~(\S\ref{sec:exp:feature-surrog}), which approximate a black-box model with an interpretable predictor across the entire feature space.
CO and formal verification methods are frequently employed in these tasks, as they enable subset- or cardinality-minimal explanations, yield exact or certified approximations, support complex domain constraints (e.g., actionability, plausibility, fairness), and provide principled tools for exploring provably faithful or diverse explanations. 

\subsection{Feature-subset explanations}
\label{sec:exp:feature-subset}

This category of methods focuses on \emph{which features matter} for a given prediction. Explanations take the form of subsets of features whose alteration (for counterfactual or contrastive explanations) or fixing (for abductive explanations) is sufficient to alter or guarantee the model’s output.

\paragraph{Counterfactual explanations.}
A counterfactual answers the question: ``Which minimal changes to this example would yield a different prediction?'' Given an instance $\mathbf{\hat{x}}$, it is defined as a closest feasible $\mathbf{x}$ that alters the model’s output, e.g., flips the predicted class \citep{Wachter2018}:
$$\mathbf{x} \in \arg\min_{\mathbf{x'} \in \mathcal{X}_{\text{feasible}}} d(\mathbf{\hat{x}},\mathbf{x'}) \quad \text{s.t. } h(\mathbf{x'}) \neq h(\mathbf{\hat{x}}),$$
where $\mathcal{X}_{\text{feasible}}$ denotes the feasible domain, and $d(\cdot,\cdot)$ is a user-specified distance capturing the notion of effort or cost of change. Beyond proximity and validity, trustworthy counterfactuals should be \emph{actionable} (modify only mutable features and respect constraints on the range of possible actions), \emph{sparse} (change as few features or take as few actions as possible), \emph{plausible} (remain within the data manifold or satisfy structural rules), and \emph{robust} (continue to flip the prediction under small perturbations or model updates). Additional desiderata such as \emph{diversity} or \emph{fairness} across explanations are also commonly required. Some recent surveys provide comprehensive taxonomies of these criteria \citep{Verma2024,Guidotti2024}.

Depending on the underlying predictive model, computing counterfactuals can range from straightforward (e.g., linear models) to computationally challenging (deep neural networks or large tree ensembles). Early, model-agnostic approaches based on gradient descent \citep[see, e.g.][]{Wachter2018,Mothilal2020,Lucic2022} offer scalability but come with well-documented drawbacks: they can converge to poor local minima and spectacularly overshoot the necessary changes \citep{Parmentier2021c}. These methods may not yield a feasible counterfactual in cases where one exists, be highly susceptible to small perturbations \citep{Slack2021}, and do not easily incorporate structural constraints that define plausible and actionable explanations, thereby exposing end users to explanations that may be invalid or unnecessarily costly. To overcome these limitations, recent work formulates counterfactual explanations as combinatorial optimization or feasibility problems (via MILP/MIQP, CP, MaxSAT, or SMT). Such formulations explicitly encode the predictive model, the distance function (or action cost), and rich structural constraints, enabling provably minimal, feasible, and consistent explanations. 

CO-based explanation methods have been particularly successful for tree ensembles due to their combinatorial and piecewise linear structure. Early explanation methods for these models include \citet{Cui2015} and \citet{Kanamori2020}, which rely on discretizations of features using binary variables. Better scalability was achieved by the OCEAN formulation of \citet{Parmentier2021c}, presented in Highlight~\ref{highlight:OCEAN}, using root‑to‑leaf continuous flow variables and binary branching choices at each depth, without the need for big-$M$ logic or discretization. This yields a tight formulation that can handle heterogeneous feature types (numerical, ordinal, binary, categorical), oblique splits, hard or soft voting, and multiclass outputs. Moreover, this framework integrates plausibility constraints via isolation forests to avoid counterfactuals that correspond to distributional outliers, and naturally accommodates actionability restrictions by encoding structural relationships and logical implications among features as constraints within the MILP.

\begin{figure*}[!htb] \refstepcounter{highlight}
\centering \scalebox{0.95}{\begin{highlight}[ycornermulticlass]{Optimal Counterfactual Explanations \citep{Parmentier2021c}}
\label{highlight:OCEAN}
\begin{subequations}
\begin{align}
    \min_{\mathbf{z}, \boldsymbol{\eta}, \mathbf{x}, \mathbf{s}} \quad & \sum_{f\in[p]} c_f (x_f - \hat{x}_f)^2 \label{eq:cf_obj} \\
    \text{s.t.} \quad &z_{t,1} = 1 & \forall t \in \mathcal{T} \label{eq:ocean_flow_init}\\
    &z_{t,j} = z_{t,\ell(j)} + z_{t,r(j)} & \forall t \in \mathcal{T}, j \in \mathcal{B}_{t}\label{eq:ocean_flow_conservation}\\
    & \sum_{j \in \mathcal{B}_{td}} z_{t,\ell(j)} \leq \eta_{td} & \forall t \in \mathcal{T}, d \in \mathcal{D}_t\label{eq:ocean_flow_integrality}\\
    & \sum_{j \in \mathcal{B}_{td}} z_{t,r(j)} \leq 1 - \eta_{td} & \forall t \in \mathcal{T}, d \in \mathcal{D}_t\label{eq:ocean_flow_integrality_bis}\\
    & x_f \leq 1 - z_{t,\ell(j)}, \  x_f \geq z_{t,r(j)}  & \forall f \in [p], t \in \mathcal{T}, j \in \mathcal{B}_{tf} \label{eq:cf_cons4} \\
    & \mathbf{x} \in \mathcal{X}_{\text{plausible}} \cap \mathcal{X}_{\text{actionable}} \label{eq:cf_cons7} \\
    & s_y = \sum_{t \in \mathcal{T}} \sum_{j \in \mathcal{L}_t} w_t p_{tjy} z_{t,j} & \forall y \in \mathcal{Y} \label{eq:ocean_cf_1} \\
    & s_{y^*} > s_y & \forall y \in \mathcal{Y} \setminus \{y^*\} \label{eq:ocean_cf_2}
    &
\end{align}
\end{subequations}
Objective~\eqref{eq:cf_obj} minimizes a Mahalanobis distance between the original instance $\mathbf{\hat{x}}$ and the counterfactual $\mathbf{x}$, with $c_f$ being a feature-specific cost value. Constraints~\eqref{eq:ocean_flow_init} initialize the flow (modeling the path of the counterfactual) at the root of each decision tree $t \in \mathcal{T}$, while Constraints~\eqref{eq:ocean_flow_conservation} ensure its propagation through each branching node $j \in \mathcal{B}_t$ of tree $t$, with $\ell(j)$ and $r(j)$ being its left and right children. Integrality of the continuous flow variables $z_{t,j} \in [0,1]$ is ensured in Constraints~\eqref{eq:ocean_flow_integrality} and~\eqref{eq:ocean_flow_integrality_bis} thanks to the binary variable $\eta_{td}$, indicating whether the counterfactual branches to the left ($\eta_{td}=1$) or to the right ($\eta_{td}=0$) at depth $d$ of tree $t$, with $\mathcal{D}_t$ denoting all the possible depths within tree $t$ and $\mathcal{B}_{td}$ the set of branching nodes at depth $d$ of tree $t$. Constraints~\eqref{eq:cf_cons4} ensure that the counterfactual's feature values are consistent with the splits along its path, where $\mathcal{B}_{tf}$ is the set of branching nodes within tree $t$ involving feature $f$. While Constraints~\eqref{eq:cf_cons4} are specific to binary features, additional extensions of this model permit dealing with continuous, ordinal, and categorical features.
Constraint~\eqref{eq:cf_cons7} ensures that the counterfactual explanations are plausible and actionable. 
Constraints~\eqref{eq:ocean_cf_1} compute the ensemble's confidence score towards each label $y \in \mathcal{Y}$, where $w_t$ is the weight of tree $t$ in the soft voting process and $p_{tjy}$ is the confidence score associated to class $y$ in leaf $j \in \mathcal{L}$ of tree $t$.
Finally, Constraints~\eqref{eq:ocean_cf_2} ensure that the target label $y^*$ yields a higher confidence score than any other, making $\mathbf{x}$ a valid counterfactual of class $y^*$.
\end{highlight}} 
\end{figure*}

Other declarative programming frameworks have also been explored. MACE \citep{Karimi2020a} encodes both the model and the distance function into SMT and solves a sequence of satisfiability problems under a binary search over the distance budget. It is model-agnostic, provided the predictor can be expressed as a quantifier-free logical program involving linear arithmetic and Boolean branching (e.g., comparisons and if-then-else statements). \citet{Raevskaya2025} developed a MaxSAT formulation for tree ensembles; however, their method is applicable only to hard-voting models. Building on this line of work, \citet{Khouna2026b} proposed a CP formulation, extended the MaxSAT encoding to support soft voting through pseudo-Boolean encodings, and systematically compared MILP, CP, SMT, and MaxSAT encodings, identifying the computational regimes in which each paradigm is most efficient. Overall, CP appears to be the most versatile, but MaxSAT excels for hard-voting tree ensembles, and MILP remains competitive in amortized inference settings with a moderate number of split levels.

Some mathematical programming algorithms have also been developed to explain other classes of models. For linear classifiers, \citet{Ustun2019b} cast actionable recourse as a MIP enforcing immutability, one-sided constraints, and feature-specific costs. For neural networks, \citet{Mohammadi2021} presented a MIP framework for nearest counterfactuals and discussed SMT-based alternatives, whereas for $k$‑nearest neighbors, \citet{Contardo2024} developed MILP and CP formulations and showed how decomposition techniques can improve tractability.

Some studies considered other desiderata that are important for counterfactual generation. \citet{Kanamori2021} focused on the generation of actionable and ordered counterfactual explanations, extending earlier work on actionable recourse \citep{Ustun2019b} and feasibility-aware paths \citep{Poyiadzi2020}. Their framework formulates the joint search over perturbations and feature-change orderings as a MILP, with costs that depend on causal or interaction structures among features. Mathematical formulations also enable global combinatorial tasks such as group-level recourse. Notably, \citet{Carrizosa2024a,Carrizosa2024} proposed MIP models for collective counterfactual explanations, aiming to generate a set of diverse explanations that efficiently cover a group of individuals, while \citet{Lodi2024} cast ``one-for-many'' recourse as a column-generation problem that seeks a small set of shared recourse actions covering many individuals. In a related direction, \citet{Chatzis2026} proposed SOGAR, which learns \emph{recourse summary trees}, i.e., shallow decision trees that partition the affected population into subgroups and assign one shared sparse action to each leaf. The method formulates this task as a bi-objective ODT problem, producing in a single run a set of cost-effectiveness trade-offs under depth, leaf-count, and action-sparsity constraints. Finally, \citet{Khouna2026} studied the interactive generation of optimal counterfactual explanations with millisecond latency by performing an offline preprocessing phase that constructs a geometric index, termed a \emph{counterfactual map}, which subsequently allows extremely efficient resolution of online queries.

\paragraph{Abductive explanations.}
In a distinct manner, an abductive explanation (AXp) answers the question: ``Which minimal facts about this example guarantee the same prediction?'' Given an instance $\mathbf{x}$, an AXp is a subset-minimal set of feature-value literals $F$ such that:
$$\forall \mathbf{x}' : \bigwedge_{f \in F} (x'_f = x_f) \;\Rightarrow\; h(\mathbf{x}') = h(\mathbf{x}).$$
Abductive explanations also subsume the concept of \emph{anchors} \citep{Ribeiro2018}, which give heuristic approximations of sufficient reasons via local perturbations and set-cover heuristics.
In propositional terms, AXps coincide with \emph{prime implicants} of the classifier restricted to the predicted class, and can be sought as subset-minimal or cardinality-minimal sets \citep{Ignatiev2019}. A general approach casts the predictor as a feasibility model in a declarative-programming framework (e.g., SAT/SMT/MILP/MaxSAT) and resolves entailment queries via feasibility checks; subset-minimal AXps then require a linear number of feasibility checks (deletion test), whereas cardinality-minimal AXps rely on hitting-set optimization. This view applies to a wide range of models, provided a mathematical model is available. For random forests, \citet{Izza2021a} showed that a purely propositional SAT encoding of paths and majority voting yields scalable AXp computation. For additive tree ensembles (e.g., boosted trees), MaxSAT can encode class-score comparisons as optimization objectives, enabling efficient AXp extraction \citep{Ignatiev2022}.


Across all these applications, mathematical programming approaches provide certificates of minimality/fidelity and a principled methodology to impose domain constraints. In other words, they make additional desiderata explicit and checkable.

\subsection{Feature-Attribution Explanations}
\label{sec:exp:feature-attrib}

This category of methods focuses on \emph{how much} each feature contributes to a prediction. Rather than identifying a subset of features that is sufficient to change or preserve the output, feature-attribution methods assign a score $\phi_f(\mathbf{x})$ to each feature $f$, measuring its contribution to the prediction $h(\mathbf{x})$ for a fixed input $\mathbf{x}$. In this sense, they decompose a prediction into feature-level contributions instead of constructing alternative inputs.

A large portion of the literature in this domain is based on Shapley values from cooperative game theory. Let $F=\{1,\dots,p\}$ denote the set of features and, for a fixed instance $\mathbf{x}$, let $v_\mathbf{x}:2^F\to\mathbb{R}$ be a \emph{value function} assigning a score to each coalition $S\subseteq F$ (e.g., an expected model output given the feature values $\mathbf{x}_S$). The Shapley value of feature~$f$ at $\mathbf{x}$ is
$$
\phi_f(\mathbf{x})
=
\sum_{S \subseteq F \setminus \{f\}}
\frac{|S|!\,(p-|S|-1)!}{p!}\,
\bigl(v_\mathbf{x}(S\cup\{f\}) - v_\mathbf{x}(S)\bigr),
$$
i.e., the average marginal contribution of $f$ over all subsets $S$ not containing it. With suitable choices of $v_\mathbf{x}$ and of a baseline prediction $\mathbb{E}[h(\mathbf{X})]$, Shapley values are the unique additive attributions satisfying local accuracy, consistency, and missingness \citep{Strumbelj2011}.
Computationally, however, exact Shapley values average marginal contributions over all $2^p$ feature coalitions and are NP-hard to compute in general \citep{Deng1994}.
To circumvent these computational barriers, KernelSHAP recasts Shapley estimation (i.e., approximation) as a weighted least-squares regression over sampled coalitions 
\citep{Lundberg2017,Covert2021}. More specifically for decision trees and tree ensembles, TreeSHAP provides a polynomial-time algorithm that leverages the path structure of trees via dynamic programming, rather than explicit enumeration \citep{Lundberg2020}, thereby computing Shapley values exactly for a specific form of value function. That said, several recent works have highlighted important limitations of Shapley-based attributions, including sensitivity to feature correlations, violations of intuitive causal relevance, and misleading importance assignments under certain model structures. The choice of value function, in particular, has a substantial impact on the resulting attributions \citep{Kumar2020a,Marques-Silva2024}.


\subsection{Prototype-based Explanations}
\label{sec:exp:feature-proto}

This category of methods focuses on \emph{which training examples best explain a decision}. Prototype-based explanations justify $h(\mathbf{x})$ by pointing to a small set of representative training instances (i.e., ``you were classified like these cases''), sometimes complemented by \emph{criticisms} designed to highlight data points where such prototypes fail to capture the data distribution.

A classical line of work formalizes prototype selection as a discrete optimization problem over training instances. 
\citet{Bien2011} assumed that each data point can serve as a prototype whose ball covers nearby examples of the same class. They cast the search for a small and accurate prototype set as a prize-collecting set cover problem (Highlight~\ref{highlight:Prototype}), trading off (i) uncovered points of the correct class, (ii) miscovered points of other classes, and (iii) the number of selected prototypes. Related prototype-selection formulations also appear in nearest-neighbor classification, where one fixes the number of prototypes and minimizes empirical misclassification over that subset \citep{Carrizosa2007a}, leading to variants of $p$‑median and facility-location problems with classification-driven costs. In MMD-critic, \citet{Kim2016} defined a kernel-based discrepancy between the empirical data distribution and a small set of prototypes, and formulated an objective that is approximately submodular in the chosen subset. Prototypes are then obtained by greedy maximization of this set function subject to a cardinality constraint, with approximation guarantees from monotone submodular maximization. Criticisms, designed to highlight parts of the data not well represented by the prototypes, are also selected by maximizing a witness function augmented with a regularizer that favors diversity.

\begin{figure*}[!htbp] \refstepcounter{highlight}
\centering \scalebox{0.95}{ \begin{highlight}[ycornermulticlass]{Prototype Selection \citep{Bien2011}}
\label{highlight:Prototype}
\begin{subequations}
\begin{align}
    \min_{\mathbf{u}, \boldsymbol{\xi}, \boldsymbol{\eta}} \quad & \sum_{i \in [n]} \xi_i + \sum_{i\in [n]} \eta_i + \lambda \sum_{j \in [n]} u_j \label{eq:pcsc_obj} \\
    \text{s.t.} \quad & \sum_{j \in [n]} u_j \mathbb{1}[y_j = y_i~\text{and}~\mathbf{x}_i \in B(\mathbf{x}_j)]  \geq 1 - \xi_i & \forall i \in [n] \label{eq:pcsc_cons1} \\
    & \sum_{j \in [n]} u_j \mathbb{1}[y_j  \neq y_i~\text{and}~\mathbf{x}_i \in B(\mathbf{x}_j)] \leq \eta_i & \forall i \in [n] \label{eq:pcsc_cons2} 
\end{align}
\end{subequations}

Objective~\eqref{eq:pcsc_obj} includes three components: (a) the number of training examples not covered by a prototype of their own class, (b) the number of examples wrongly covered by prototypes of other classes (each penalized by the number of prototypes wrongly covering them), and (c) the number of selected prototypes, weighted by a regularization parameter $\lambda$. 
$B(\mathbf{x}_j)$~is defined as the ball with radius $\epsilon$ centered around prototype $\mathbf{x}_j$, such that any example $\mathbf{x}_i \in B(\mathbf{x}_j)$ is covered by prototype $\mathbf{x}_j$.
Constraints~\eqref{eq:pcsc_cons1} ensure that each training example $\mathbf{x}_i$ is covered by at least one prototype of its own class unless the slack variable $\xi_i = 1$. Constraints~\eqref{eq:pcsc_cons2} ensure that no training example $\mathbf{x}_i$ is wrongly covered by a prototype from another class, unless $\eta_i > 0$. Finally, the binary decision variables $u_j$ determine whether $\mathbf{x}_j$ is selected as a prototype, and variables $\xi_i$ and $\eta_i$ are non-negative.
\end{highlight}} \end{figure*}

Prototypes are also closely connected with interpretable clustering.
\citet{Carrizosa2022} tackled the problem of explaining the output of a clustering algorithm by selecting, for each cluster, one or several prototypes such that cluster members are close to their prototype and nonmembers are not. This leads to a biobjective MILP inspired by covering-location problems, maximizing the total number of true positives (cluster points well covered by their prototypes) while minimizing false positives (points in other clusters that fall within a prototype’s influence). \citet{Lawless2023} proposed an alternative cluster-explanation framework in which each cluster is described by a sparse polyhedron. They formulated an integer program with an exponential number of candidate half-spaces and solved it via column generation. Other works on explainable or interpretable clustering focused on tree-based cluster descriptions, reducing them to variants of $k$-means or set-covering problems \citep[see, e.g.,][]{Dasgupta2020,Bertsimas2021}.

\subsection{Surrogate Models}
\label{sec:exp:feature-surrog}

This category of methods focuses on \emph{learning a simpler model that mimics a complex predictor}. Given a ``teacher'' $h$, a surrogate $g \in \mathcal{G}$ (e.g., a decision tree, rule list, or sparse linear model) is trained to approximate $h$ while remaining intrinsically interpretable. 
In contrast to feature-subset or prototype-based explanations, which explain a given prediction in terms of input features or training examples, surrogate models aim to provide a \emph{model-level} abstraction: a user can reason directly with $g$ instead of querying $h$. Moreover, unlike model pruning, which simplifies a model within its own parametric class, surrogates can belong to a different, explicitly interpretable hypothesis class.
A generic formulation is:
\begin{equation}
\min_{g \in \mathcal{G}} 
L(g,h) 
+ \lambda\, R(g),
\end{equation}
where $L(g,h)$ measures \emph{fidelity} to the teacher and the regularizer $R(g)$ enforces sparsity or other structural desiderata.

\paragraph{Empirical surrogates.}
Most common surrogates are approximate, requiring agreement with the teacher only on a finite set of instances. Classic examples include local linear surrogates such as LIME \citep{Ribeiro2016}, early rule- or tree-extraction methods for neural networks such as TREPAN \citep{Craven1996}, and subsequent model distillation approaches \citep[e.g.,][]{Hinton2015,Frosst2018a} where one trains an interpretable classifier on pseudo-labels $h(\mathbf{x}_i)$. To that end, MILP, CP, and MaxSAT frameworks used for optimal decision trees, rule lists, or decision sets (see Section~\ref{sec:train-interp}) could be directly reused by replacing ground-truth labels with the teacher's outputs, yielding surrogates that optimally trade fidelity and interpretability under additional structural constraints (e.g., sparsity, bounded depth, monotonicity).

\paragraph{Faithful surrogates.}
At the opposite end of the spectrum lie faithful surrogates, which aim to reproduce the teacher exactly over the feature space (or parts thereof). The fidelity requirement responds to the broader argument expressed in \citet{Rudin2019}, who cautions that approximate surrogates introduce an additional layer of opacity and potential error. 
In contrast to empirical fidelity, faithful surrogate methods enforce \emph{functional (predictive) equivalence} constraints of the form $g(\mathbf{x}) = h(\mathbf{x})$ for all $\mathbf{x}$ in the domain of interest. For tree ensembles, \citet{Vidal2020a} showed that any boosted or random-forest model can be compiled into a single decision tree of minimal size that is functionally identical to the ensemble.
Their ``born-again trees'' algorithm solves this NP-hard search problem using DP over the regions induced by the ensemble, along with bounding rules that guarantee minimality of the resulting tree. This contrasts with MILP-based formulations that encode functional equivalence through exponentially many constraints and rely on cutting-plane separation (as in \citealt{Emine2025} for faithful ensemble pruning).

Importantly, the computational difficulty of enforcing functional equivalence in surrogate models depends strongly on the hypothesis classes considered. For example, functional equivalence between two decision trees can be tested in polynomial time by simple inspection of the leaves. In contrast, equivalence tests involving tree ensembles are substantially harder: \citet{Vidal2020a} demonstrated that even deciding whether a single decision tree is functionally equivalent to a given tree ensemble is NP-hard.

Moreover, between complete fidelity over the entire feature space and empirical fidelity on a finite set of points, there are various alternatives. For example, one can impose functional identity requirements only on inputs or regions of the feature space that are \emph{plausible}, such as regions that lie on the data manifold, satisfy density or structural criteria, or are constrained to avoid outliers. Such restrictions also allow focusing the search on regions of interest, producing more local explanations. Born-again trees can support this relaxation by limiting the DP reconstruction to a local subregion (hyperrectangle) or to subregions that contain data points, similar to \cite{Parmentier2021c} and \cite{Carreira-Perpinan2023}, which restrict the search for counterfactual explanations to plausible or live regions, respectively.

\paragraph{Hybrid and partially interpretable models.}
Surrogate modeling is also closely related to \emph{hybrid} or \emph{partially interpretable} models, which couple an interpretable component $h_{\mathrm{I}}$ with a black-box component $h_{\mathrm{BB}}$ through an explicit routing mechanism. A typical design produces
\begin{equation}
h(\mathbf{x}) \;=\;
\begin{cases}
h_{\mathrm{I}}(\mathbf{x}), & \text{if } \mathbf{x} \in \Omega,\\
h_{\mathrm{BB}}(\mathbf{x}), & \text{otherwise},
\end{cases}
\end{equation}
where $\Omega$ is an interpretable region of the feature space. Hybrid predictive models \citep{Wang2021} jointly learn $h_{\mathrm{I}}$ (e.g., a rule set or rule list) and the coverage region $\Omega$ using a multi-objective criterion balancing predictive accuracy, model complexity, and \emph{transparency}, defined as the fraction of instances handled by the interpretable component.

Early work by \citet{Pan2020} introduced \emph{interpretable companions}, rule-list surrogates designed to approximate a black-box predictor on an interpretable subset of the domain. Companions are obtained by solving a discrete optimization problem over rule lists through local search, optimizing the area under the transparency-accuracy curve. Later on, \citet{Ferry2024a} provided a unified treatment of hybrid interpretable models and derived statistical generalization guarantees. Building on CORELS’ B\&B procedure for learning optimal rule lists, their algorithms embed an optimal rule-list solver within the hybrid learning pipeline, yielding a certificate of optimality for the interpretable component and explicit control over transparency. Like prior work, their first method trains the interpretable component (and its coverage region) on top of a pre-trained black-box model. They also proposed a second method that reverses this order: it first fits the interpretable component, then trains a black-box specialized to the subspace it will ultimately handle.

\section{CO for ML Fairness}

Fairness in ML aims to prevent disparities in performance or outcomes across individuals or groups defined by sensitive attributes (e.g., sex, race). A standard distinction is between \emph{group fairness} and \emph{individual fairness} \citep{Verma2018}. Group fairness requires parity (exact or approximate) of statistical criteria across protected groups, such as \emph{demographic parity} (selection rates)~\citep{Dwork2012}, \emph{equal opportunity} (true positive rates), \emph{equalized odds} (both false positive and false negative rates)~\citep{Hardt2016} or \emph{predictive parity} (positive predictive values)~\citep{Chouldechova2017}. Individual fairness~\citep{Dwork2012} emphasizes that models should deliver consistent decisions for \emph{similar} individuals, i.e., those with nearly identical profiles except for the sensitive attribute.

A key practical challenge is that these notions are not simultaneously satisfiable in general. When base rates differ across groups, classical results show that several desirable properties (e.g., predictive parity and error-rate parity) cannot hold simultaneously, making fairness inherently multi-criteria \citep{Kleinberg2017, Chouldechova2017}.

\subsection{Training Inherently Fair Models}

Fairness-enhancement methods can be broadly classified into three main approaches: (i) preprocessing, which focuses on transforming the input data to improve fairness, (ii) in-processing, which involves adapting the model or the learning algorithm to integrate fairness aspects, and (iii) postprocessing, which flags unfair responses or modifies the output to improve the level of fairness \citep[see, e.g.][]{Caton2024}. We primarily focus on in-processing techniques, which most naturally give rise to CO subproblems and generally offer the best fairness-accuracy trade-offs~\citep{Barocas2023}. We then discuss preprocessing and postprocessing methods.

\subsubsection{Fair Learning Approaches}
\paragraph{Group-based statistical fairness constraints.}
Several of the models discussed in Section~\ref{sec:Transparency} can be extended to incorporate group fairness constraints or fairness-aware objective terms. In conventional gradient-based learning, those constraints are typically enforced through Lagrangian relaxations or surrogate penalty terms \citep{Agarwal2018,Cotter2019}. Moreover, notable examples of mathematical programming models for supervised learning that have been extended to ``fair versions'' include \citet{Aghaei2019a,VanderLinden2022} and \citet{Jo2023}, which proposed MILP and DP approaches for learning optimal and fair decision trees. For rule lists, \citet{Aivodji2019a,Avodji2021} introduced FairCORELS, a multiobjective extension of CORELS that uses B\&B to enumerate accuracy-fairness trade-offs under statistical fairness metrics. In a similar vein, \citet{Deza2024} proposed a flow-based formulation for fair regression, whereas \citet{Lawless2023a} and \citet{Rober2025} relied on column generation and integer programming to construct fair rule-based models, highlighting how fairness constraints interact with sparsity and interpretability. While the majority of research on fairness in machine learning has focused on binary classification, \citet{Rouzot2022} extended several fairness metrics to the multi-class setting and proposed a MILP formulation for learning optimal fair scoring systems.

Exact count-based fairness constraints are also generally challenging, as the \emph{counts} of classified or misclassified samples across subgroups require specific integer variables and logical constraints in mathematical programs. To address this, \citet{Zafar2017a,Zafar2019} proposed convex proxy constraints for linear classifiers, based on bounding the \emph{covariance} between the sensitive attribute and the signed prediction score. Optimizing or constraining this quantity encourages behavior similar to demographic parity while yielding tractable convex formulations for logistic regression and SVMs (see Highlight~\ref{highlight:Fair}). These formulations also support the notion of \emph{business necessity}, in which one maximizes fairness subject to a lower bound on predictive performance. More generally, balancing fairness, sparsity, and accuracy naturally leads to multi-objective optimization problems, whose mathematical formulations enable the systematic exploration of achievable trade-offs and admissible performance ranges \citep{Langlade2025}.

\begin{figure*}[!htbp] \refstepcounter{highlight}
\centering \scalebox{0.95}{ \begin{highlight}[ycornerbinary]{Learning Fair Logistic Regression Models~\citep{Zafar2019}}
\vspace*{0.05cm}
\label{highlight:Fair}
\begin{subequations}
\begin{align}
    \min_{\mathbf{w}} \quad & \sum_{i \in [n]} \log \left(1 + e^{-y_i \mathbf{w}^\top \mathbf{x}_i} \right) \label{eq:logistic_obj} \\
    \text{s.t.} \quad & \left| \frac{1}{n} \sum_{i \in [n]} (s_i - \bar{s}) \mathbf{w}^\top \mathbf{x}_i \right| \leq c \label{eq:disparate_impact_cons}
\end{align}
\end{subequations}

Objective~\eqref{eq:logistic_obj} minimizes the logistic loss, where $\mathbf{w}$ are the model parameters. The fairness constraint~\eqref{eq:disparate_impact_cons} limits the covariance between the sensitive attribute $s$ and the decision boundary $\mathbf{w}^\top \mathbf{x}$, thereby providing a tractable proxy for controlling demographic parity (or disparate impact). More precisely, letting $s_i$ be the sensitive attribute value of example $i$ and $\bar{s}$ the average sensitive attribute value across the training dataset, the empirical covariance term is given by $\frac{1}{n} \sum_{i \in [n]} (s_i - \bar{s}) \mathbf{w}^\top \mathbf{x}_i$. Bounding this quantity yields a tractable, convex model amenable to gradient-based optimization.
\end{highlight}} \end{figure*}

\paragraph{Intersectional constraints.}
The aforementioned approaches typically assume a fixed, small set of protected groups, for which fairness constraints can be explicitly encoded. More generally, enforcing fairness across a combinatorial family of \emph{intersectional} subgroups is formulated as a multicalibration problem by \citet{Hebert-Johnson2018}, and addressed through cut generation in \citet{Kearns2018} and \citet{Nemecek2026}. In these works, a master problem optimizes accuracy subject to a current subset of fairness constraints, while a separation oracle iteratively identifies and reintroduces the most violated subgroup constraint, until a model is obtained that satisfies fairness over all considered intersections.

\paragraph{Fair clustering.}
Group fairness considerations have also been studied in some unsupervised learning settings. In fair clustering, \citet{Chierichetti2017} introduced the \emph{fairlet} decomposition as a min-cost flow solution to achieve fair pre-aggregation, followed by standard clustering. Later work improved scalability and generality, e.g., near-linear-time fairlet computation \citep{Backurs2019a} and approximation algorithms that handle multiple protected groups and $\ell_p$ objectives \citep{Bera2019}. More recently, \citet{Lawless2024} proposed a single-phase MILP formulation, along with decomposition techniques, for fair clustering based on \emph{minimum representation} constraints, and showed that even assigning points to fixed centers becomes NP-hard under fairness requirements. A variety of other works addressed related problems stemming from connections to gerrymandering in the context of political districting.

\paragraph{Individual fairness.}
In contrast to group-based fairness notions, which impose statistical conditions over aggregates, \emph{individual fairness} requires that similar individuals receive similar predictions. The original formulation of \citet{Dwork2012} imposes a global Lipschitz condition
$
\| h(\mathbf{x}) - h(\mathbf{x'}) \|
\;\le\;
L\, d(\mathbf{x},\mathbf{x'})$
for all $\mathbf{x}, \mathbf{x'}$,
where $d(\cdot,\cdot)$ is a similarity metric and $L$ is a Lipschitz constant. In practice, this constraint is often relaxed into a local robustness-style condition:
$$
\| h(\mathbf{x}) - h(\mathbf{x'}) \|
\;\le\;
\tau
\quad
\forall \mathbf{x},\mathbf{x'}:\, d(\mathbf{x},\mathbf{x'}) \le \epsilon,
$$
which bounds prediction variation within a neighborhood of similar points (obtained by local perturbations or changes in protected attributes). Enforcing such pairwise constraints directly is generally intractable; therefore, most approaches incorporate relaxed surrogates into the training objective.

In \citet{Yurochkin2020}, similarity is defined via a specific distance (e.g., a Mahalanobis metric that downweights or ignores sensitive directions), and training minimizes a Wasserstein distributionally robust objective over a ball around the empirical distribution, with transport cost induced by the fair metric. This encourages the model to remain accurate under worst-case perturbations of the data along directions that should not affect the prediction, such as sensitive demographic directions. An established relaxation is \emph{distributional individual fairness} \citep{Yurochkin2021}. Let $Q$ denote the data distribution and $d(\mathbf{x},\mathbf{x'})$ the dissimilarity metric. It requires:
$$
\sup_{\pi \in \Pi(Q,Q)}
\; \mathbb{E}_{(\mathbf{x},\mathbf{x'})\sim \pi}
\big[\, \|h(\mathbf{x}) - h(\mathbf{x'})\| \,\big]
\;\le\; \tau
\quad
\text{s.t.}
\quad
\mathbb{E}_{(\mathbf{x},\mathbf{x'})\sim \pi}
\big[\, d(\mathbf{x},\mathbf{x'}) \,\big]
\le \epsilon,
$$
where $\Pi(Q,Q)$ denotes the set of couplings of $Q$ with itself. This condition replaces pointwise constraints with a worst-case transport coupling. \citet{Yurochkin2021} derived a dual form of this constraint, yielding a tractable regularizer that is integrated into the learning loss.

Complementary to these approaches, LCIFR \citep{Ruoss2020} learns a representation that maps each instance and its similar counterparts into a small latent neighborhood, reducing fairness to a certifiable local robustness property of the downstream classifier, based on the DL2 training framework \citep{Fischer2019}. CertiFair \citep{Khedr2023} instead constructs paired networks and derives fairness losses using neural-network bounding procedures. Finally, \citet{Ehyaei2024} linked individual fairness to Wasserstein distributional robustness under structural causal models.

\subsubsection{Preprocessing and Postprocessing Approaches}

\paragraph{Preprocessing.} 
Preprocessing methods modify the training data (sensitive attributes, non-sensitive attributes, or labels) to remove undesired correlations while preserving as much information as possible~\citep{Caton2024}. A first line of work learns fair intermediate representations. In a seminal contribution, \citet{Zemel2013} mapped each example to a distribution over prototypes and optimized a nonlinear objective balancing demographic parity, reconstruction, and prediction, using L-BFGS. Building on this fair-representation paradigm, \citet{Song2019} formulated controllable fair representation learning as a constrained information-theoretic problem, interpreting earlier methods, including \citet{Zemel2013}, as approximations of a Lagrangian dual with fixed trade-off weights. Although the idealized problem over distributions enjoys strong duality, the practical parameterized training problem remains nonlinear and non-convex and is solved using alternating gradient-based updates. Similarly, \citet{Lahoti2019} proposed iFair, a low-rank probabilistic mapping model optimized with L-BFGS, whose objective balances utility and individual fairness.

A second line of work directly transforms or reweights the data distribution.
\citet{Calmon2017} proposed learning a randomized data transformation by minimizing a distributional utility loss subject to discrimination-control and individual-distortion constraints, and showed that, under suitable choices, the resulting problem is convex or quasiconvex. Related approaches rely on optimal transport, either to repair the input data through Wasserstein barycenters~\citep{DelBarrio2019} or to transform group-specific score distributions toward a common intermediate distribution~\citep{Zehlike2020}.

Closer to CO, \citet{Xiong2024} introduced FairWASP, a MILP formulation for finding integer weights such that the reweighted dataset remains close to the original one in Wasserstein distance while satisfying demographic parity constraints (Highlight~\ref{highlight:FairWASP}). They further showed that optimal solutions to the integer-weighting problem are also optimal for its linear relaxation, and designed an efficient cutting-plane algorithm. In a follow-up work, \citet{Xiong2024b} extended this perspective to fair Wasserstein \emph{coresets}, i.e., weighted synthetic representative samples that remain close to the original data distribution while satisfying demographic parity constraints. Their method reformulates the problem as a nested minimization problem with linear fairness constraints and solves it using a dedicated majority-minimization procedure built on a cutting-plane variant of FairWASP. Finally, \citet{Burgard2026} addressed fair and calibrated synthetic data generation rather than dataset repair. They cast the assignment of labels to unlabeled synthetic samples as a MILP with fairness constraints, showed that a unimodularity property makes the linear relaxation sufficient, and introduced a preprocessing step to reduce computational cost.

\begin{figure*}[!htb] \refstepcounter{highlight}
\centering \scalebox{0.95}{ \begin{highlight}[ycornermulticlass]{Fair Wasserstein Preprocessing \citep{Xiong2024}}
\label{highlight:FairWASP}
\begin{subequations}
\begin{align}
    \min_{\mathbf{w} \in \mathbb{N}_0^n} \quad & \mathcal{W}_M(\hat{p}_{\mathbf{1}_n}, \hat{p}_{\mathbf{w}}) \label{eq:fairwasp_obj} \\
    \text{s.t.} \quad 
    & \sum_{\substack{i=1\\ \text{s.t.}~y_i = y\\ \text{and}~s_i = s}}^{n} w_i \leq (1 + \epsilon) \frac{\sum_{i=1}^n{\mathbb{1}[y_i = y]}}{n} \sum_{\substack{i=1\\ \text{s.t.}~s_i = s}}^{n} w_i
    && \forall y \in \mathcal{Y}, s \in \mathcal{S}\label{eq:fairwasp_fairness_constr_1}\\
    & \sum_{\substack{i=1\\ \text{s.t.}~y_i = y\\ \text{and}~s_i = s}}^{n} w_i \geq \frac{1}{1 + \epsilon} \frac{\sum_{i=1}^n{\mathbb{1}[y_i = y]}}{n} \sum_{\substack{i=1\\ \text{s.t.}~s_i = s}}^{n} w_i
    && \forall y \in \mathcal{Y}, s \in \mathcal{S}\label{eq:fairwasp_fairness_constr_2}\\
    & \sum_{i=1}^n w_i = n \label{eq:fairwasp_valid_weights}
\end{align}
\end{subequations}

Objective~\eqref{eq:fairwasp_obj} minimizes the Wasserstein distance $\mathcal{W}_M$ between the empirical distribution~$\hat{p}_{\mathbf{1}_n}$ of the original dataset, in which each example $i \in [n]$ appears exactly once, and the reweighted empirical distribution $\hat{p}_{\mathbf{w}}$, in which example $i$ appears $w_i$ times. Here, $M \in \mathbb{R}^{n \times n}$ is the matrix of transport costs, where $M_{ij}$ denotes the cost of transporting mass from example $i$ to example $j$.
For each outcome value $y \in \mathcal{Y}$ and sensitive attribute value $s \in \mathcal{S}$, Constraints~\eqref{eq:fairwasp_fairness_constr_1} and~\eqref{eq:fairwasp_fairness_constr_2} enforce that the conditional probability of outcome $y$ within group $s$ under the reweighted distribution remains within a multiplicative factor $1+\epsilon$ of the marginal probability of $y$ in the (entire) original dataset. This formulation corresponds to the demographic parity fairness metric, with hyperparameter $\epsilon$ controlling the desired level of fairness.
Finally, Constraint~\eqref{eq:fairwasp_valid_weights} ensures that the new integer weights sum to $n$, so that the total mass of the reweighted dataset matches that of the original dataset.
\end{highlight}} 
\end{figure*}

\paragraph{Output-level postprocessing.} 
Output-level postprocessing modifies the predictions of an already-trained predictor, without changing the training data or the model parameters. A seminal work in this direction is \citet{Hardt2016}, who showed how to derive a fair classifier from any score-based predictor through randomized thresholding, enforcing equal opportunity or equalized odds via an LP over the induced error rates.

Several follow-up works retain this optimization-based perspective. \citet{Awasthi2020} studied the robustness of equalized-odds postprocessing when protected-group information is imperfect. 
Focusing on demographic parity, \citet{Xian2023} characterized the optimal fair postprocessing rule in the general multi-group, multi-class, and noisy setting via a Wasserstein barycenter formulation, which can be implemented as an LP. More recently, \citet{Xian2024} proposed LinearPost, a unified framework covering statistical parity, equal opportunity, and equalized odds in multiclass, attribute-aware, and attribute-blind settings.

Optimal transport and convex optimization methods provide another route to output-level postprocessing. \citet{Jiang2020} proposed mapping group-specific classifier scores to a common Wasserstein barycenter. In the multi-class setting, \citet{Denis2024} formulated fair postprocessing under demographic parity as a convex optimization problem that computes class-dependent score shifts from estimated class probabilities. They used a softmax approximation of the maximum term and solved the resulting problem via sequential quadratic programming.

Output-level postprocessing has also been studied for individual fairness. \citet{Kim2018} considered a setting in which the similarity metric is unknown and accessible only through a bounded-query comparison oracle, and proposed a convex constrained optimization framework for postprocessing the predictions of a pre-trained model. More recently, \citet{Mohammadi2023} built on the MILP modeling of ReLU neural networks proposed by \citet{Tjeng2019} (see Highlight~\ref{highlight:relu-formulations} in Section~\ref{sec:adversarial}) to introduce a fairness-verification oracle that, at inference time, identifies counterexamples differing from a query only in their sensitive attributes. Although this oracle falls within fairness verification (Section~\ref{sec:fairness:auditing}), its counterexamples are additionally used in a postprocessing step that may flip the model's prediction, ensuring that individuals sharing the same non-sensitive features receive the same outcome. The authors also proposed generating maximum-violation fairness counterexamples for sampled training examples and adding them back to the training set with the same label, thereby improving individual fairness empirically.

\paragraph{Model-level postprocessing.} 
In addition to modifying output scores or labels, some methods instead edit the trained model itself. For decision trees, FADE~\citep{Kanamori2021b} transforms a biased tree into a fairer one by solving a MILP that minimizes discrepancy from the original model, either in terms of predictions or tree structure, subject to fairness constraints. The allowed edits include deleting nodes, modifying internal branching rules, and relabeling leaves. Along the same lines, FairRepair~\citep{Zhang2020} uses MaxSMT to modify selected paths in decision trees and random forests, either by flipping their associated decisions or by inserting new splits to refine the corresponding hypercube, to satisfy fairness requirements while maintaining bounded semantic change.

Beyond tree-based models, model-level postprocessing has also been explored for neural networks and large language models (LLMs). \citet{Dai2025} studied fairness-aware neural network pruning, formulating the joint learning of sparse masks and fair weights as a constrained bilevel optimization problem. Although the pruning mask is discrete, the method relies on continuous relaxations and gradient-based updates rather than exact combinatorial optimization. \citet{Ma2025} proposed PROF, a provable fairness-repair framework that modifies a trained deep neural network in two stages: it first calibrates the feature extractor, i.e., all layers except the last one, through interval-bound-guided gradient descent to reduce feature differences and promote feature consistency across similar inputs; it then repairs the final classification layer by solving a MILP over parameter changes. Finally, \citet{Dasu2026} cast fairness-aware attention-head pruning in LLMs as a discrete search problem over binary pruning masks. Due to the high dimensionality of the search space, they tackled it with surrogate-assisted simulated annealing, using small neural surrogates to predict the bias and perplexity of each mask without repeatedly evaluating the full model.

\subsection{Fairness auditing and verification}
\label{sec:fairness:auditing}

In many applications, fairness requirements must be \emph{audited} or \emph{verified} for an already-trained model. Auditing boils down to an optimization problem: find a \emph{witness} (e.g., an input pair, a constrained region of the input domain, or a subgroup of interest) that violates a fairness criterion as strongly as possible. Verification, in contrast, aims to prove that no such witness exists within a specified domain (e.g., a similarity ball, a feasible input schema, or a data-generating model). Due to the scale of modern predictors, auditing methods are typically heuristic (e.g., test generation and guided exploration), whereas verification relies on formal or systematic approaches (e.g., SAT/SMT/MILP) that can produce \emph{certificates} but typically face size limitations. We refer to \citet{Chen2024} for a comprehensive survey and taxonomy of fairness testing.

\paragraph{Fairness audits.}
In addition to random sampling, many auditing techniques are inspired by heuristic or exact optimization strategies. Test-generation tools such as Themis \citep{Galhotra2017} quantify discrimination by sampling the input space and mutating sensitive attributes, while other extensions find discriminatory inputs using multi-start local search (e.g., two-phase global sampling followed by local perturbations in Aequitas \citep{Udeshi2018}, or population metaheuristics \citep{Fan2022}). In \citet{Sharma2021}, the search for discriminatory inputs exploits SMT models on a more tractable surrogate of the predictor. Finally, auditing can be distribution-aware: \citet{Black2020} computed counterparts across protected groups via an optimal transport problem and returned \emph{flipsets}, i.e., individuals whose outcomes differ from those of their transported counterparts.

\paragraph{Individual fairness verification.}
Individual fairness verification asks whether there exists a pair of \emph{similar} individuals whose outcomes differ, i.e., whether one can find $(\mathbf{x},\mathbf{x'})$ satisfying a similarity constraint such that $h(\mathbf{x})$ and $h(\mathbf{x'})$ disagree beyond a tolerance. This task closely connects to counterfactual search (Section~\ref{sec:exp:feature-subset}) and robustness verification (Section~\ref{sec:adversarial}). For neural networks, one can duplicate the network on two inputs, constrained to match on non-sensitive features (or remain within a similarity ball), and ask whether outputs can differ by more than a tolerance \citep{John2020,Khedr2023,Mohammadi2023}. This leads to an instance of the formal \emph{neural network verification} problem covered in detail in Section~\ref{sec:adversarial}. To that end, a rich set of methods has been developed, notably based on MILP, SMT, and B\&B search with tailored bound-propagation strategies \citep[see, e.g.,][]{Wang2021c,Duong2024b,Wu2024}. Finally, for distributional individual fairness, certification becomes a distributionally robust verification problem, where one seeks to bound the probability that a randomly drawn neighbor violates fairness under a Wasserstein-type ambiguity set \citep{Wicker2023}.

\paragraph{Group fairness verification.}
Group-fairness criteria (e.g., demographic parity, equal opportunity, equalized odds) are typically defined with respect to an input distribution (i.e., a population model) rather than a fixed test set, so verification requires reasoning about conditional probabilities of the form $\Pr[h(X)=1 \mid S=s]$ (where $X$ is a random variable for input features and $S$ for sensitive features) and about differences and ratios across groups. FairSquare \citep{Albarghouthi2017} casts these fairness statements as probabilistic model specifications and reduces their evaluation to weighted volume computation over logical encodings of the predictor. Specifically, it utilizes an SMT solver to iteratively construct disjoint hyperrectangular under-approximations of the satisfaction region, yielding monotonically improving lower and upper bounds that can permit fairness or unfairness certification. In contrast, sampling approaches, such as VeriFair \citep{Bastani2019}, treat the classifier and population model as black boxes that can be sampled, relying on adaptive concentration inequalities to evaluate fairness with high-probability guarantees, thereby yielding a probabilistic yet scalable verifier. Finally, Justicia \citep{Ghosh2021} encodes the verification of group-fairness metrics as a stochastic Boolean satisfiability (SSAT) problem for distribution-aware verification with explicit error bounds, and \citet{Boetius2025} developed a probabilistic B\&B for neural networks, computing lower/upper bounds on group-fairness probabilities by recursively splitting the input domain.

\section{CO for ML Robustness}   

Robustness in machine learning refers to the stability of predictors under perturbations of the input, the training data, or even the learned model itself. Such perturbations may be incidental (noise, measurement error, missing values) or adversarial (worst-case manipulations). Robustness desiderata lead to \emph{worst-case} formulations and typically translate into min-max optimization problems.

\subsection{Training Robust Models}
\label{sec:training-robust}

Robust training can be broadly organized according to \emph{where} perturbations occur:
(i) at training time, through outliers or label noise; 
(ii) at test time, through input perturbations around each query point; and 
(iii) at the distribution level, through shifts between the training and deployment populations.

\paragraph{Robustness to outliers and noisy labels.}
A classical approach in robust statistics is to down-weight or exclude a small fraction of corrupted observations during training. This can be implemented as a preprocessing step or integrated directly into the learning problem. In regression, the integrated approach yields trimmed estimators such as least trimmed squares (LTS), least median of squares (LMS), and least quantile of squares (LQS) \citep{Rousseeuw1984,Rousseeuw1985}.
These estimators, along with various others, form a rich family of NP-hard subset-selection problems \citep{Bernholt2006}, leading to \emph{min-min} MIPs that jointly trim outliers using binary variables and estimate the model on the retained data \citep{Giloni2002,Zioutas2009,Bertsimas2014a,Insolia2022,Puerto2025}. Yet, classical big-$M$ approaches for trimming yield extremely weak relaxations. To overcome this issue, \citet{Gomez2026} derived second-order cone reformulations that exploit the coupling between trimming decisions and strictly convex regularization terms, achieving substantial computational improvements. Related conic and bilevel formulations have also been proposed for least-quantile estimators and other structured regression problems, including exact and approximation methods for LQS and LMS estimators \citep{Bertsimas2014a,Puerto2025}.

Beyond regression, related trimming-estimation models appear across a wide range of learning tasks. In classification under training contamination, they underpin SVM formulations that penalize or budget label flips \citep{Blanco2022a}, as well as robust optimal classification trees that jointly learn splits and identify mislabeled samples \citep{Blanco2022}.
Related ideas also appear in robust covariance and shape estimation \citep{Woodruff1994}, in robust clustering where outlier selection is embedded into the clustering task \citep{Garcia-Escudero2008,Chawla2013,Gupta2017}, as well as in time-series estimation under the presence of anomalous or corrupted observations \citep{Gomez2021}. 

In parallel, robustness can also be improved through \emph{nonconvex losses} that cap the influence of extreme observations instead of nullifying them, such as ramp or hard-margin losses in SVMs and related classifiers \citep{Huang2014}. However, these losses are non-convex, and exact training becomes difficult even for linear separators, motivating the use of formulation-strengthening and bound-tightening techniques \citep{Belotti2016,Baldomero-Naranjo2020a}, decomposition or Benders-type reformulations \citep{Santana2022}, or different branching strategies \citep{Belotti2024}, to improve scalability.

\paragraph{Robustness to test-time input perturbations.}
Adversarial (or robust) training seeks a predictor that performs well under worst-case perturbations $\mathbf{x}\mapsto \mathbf{x}+\boldsymbol{\delta}$ within an uncertainty set $\boldsymbol{\delta}\in\mathcal{U}$ \citep{Madry2018}. Such a pessimistic viewpoint typically
yields \emph{min-max} optimization problems in which the learner selects a model while an adversary chooses a worst-case perturbation for each sample. For certain simple hypothesis classes, the resulting robust counterparts can be trained in closed form or reduced to tractable convex programs, and in several cases robustness to bounded input perturbations is equivalent to regularization \citep[e.g.,][]{Xu2009}. 

For discrete and structured models, robustness to input perturbations has been incorporated into combinatorial learning problems. 
\citet{Bertsimas2019e} extended robust optimization ideas to optimal decision trees and highlighted the limitations of greedy split-based heuristics in the robust setting. Subsequently, \citet{Vos2022} showed that, for decision trees with $0$-$1$ loss, the min-max adversarial training problem can be reformulated as a single-level optimization problem by characterizing the set of leaves reachable under perturbations for each sample, yielding exact MILP and MaxSAT formulations for robust optimal classification trees. Additionally, \citet{Vos2023} showed that the robustness of an existing decision tree can be improved post hoc via optimal leaf relabeling, which reduces to a minimum vertex cover problem in a bipartite graph. These exact approaches complement more scalable heuristic methods for robust training of decision trees and ensembles \citep{Vos2021a,Guo2022}.

\paragraph{Distributional robustness.}
In contrast, robustness to distribution shift concerns uncertainty in the data-generating distribution itself: rather than perturbing each sample independently, one seeks predictors whose risk remains small under a family of plausible test distributions. A standard formalization is distributionally robust optimization (DRO), which replaces empirical risk minimization by worst-case expected loss over an ambiguity set $\mathcal{P}$ of distributions around a nominal law (typically the empirical data distribution). Given a loss $\ell(\mathbf{w};\mathcal{D})$ for parameters $\mathbf{w} \in \mathcal{W}$ and data $\mathcal{D}$, DRO takes the form:
$$
\min_{\mathbf{w} \in \mathcal{W}}\ \sup_{Q \in \mathcal{P}}\ \mathbb{E}_{\mathcal{D} \sim Q}\!\left[\ell(\mathbf{w};\mathcal{D})\right].
$$
In this model, the inner maximization in DRO couples decisions across samples, preventing per-example decompositions of the adversary’s actions. Nevertheless, for many ambiguity sets, the inner optimization can admit an explicit and tractable dual reformulation.

For margin-based classifiers such as SVMs, various DRO models have been explored. A first classical line of work relies on moment-based ambiguity and distributionally robust chance constraints. \citet{Lanckriet2003} introduced the minimax probability machine, and subsequent extensions have shown that controlling worst-case misclassification probabilities under unknown distributions with prescribed moments yields tractable convex reformulations, typically second-order cone programs or semidefinite programs \citep{Shivaswamy2006,Ben-Tal2011}. Bernstein-type relaxations have also been explored, yielding better bounds while preserving tractability \citep{Ben-Tal2011}. Similar ideas also extend to kernel-based classifiers by expressing moment constraints in feature space \citep{Lin2024}.

A second line of work adopts distance-based ambiguity sets, notably Wasserstein balls, which provide a geometric notion of distributional shift and connect DRO to regularization and margin maximization. For smooth convex losses, Wasserstein DRO admits finite-dimensional convex reformulations relying on duality \citep{Shafieezadeh-Abadeh2015,Shafieezadeh-Abadeh2019,Kuhn2019}. When features are mixed or categorical, the resulting models may entail an exponential number of constraints, necessitating column-and-constraint generation or graph-based reformulations \citep{Selvi2022,Sun2025}. \citet{Singha2018} also proposed a Lagrangian relaxation of Wasserstein DRO that yields efficient training procedures along with certificates on the worst-case population objective, whereas \citet{Ho-Nguyen2023} showed that Wasserstein distributional robustness for linear classification is equivalent to minimizing a regularized ramp-loss objective, thereby connecting worst-case distributional training with conditional-value-at-risk-type margin criteria and robustness to outliers.

Finally, for discrete and structured predictors such as decision trees, the DRO inner problem cannot be dualized away. \citet{Justin2023} studied optimal classification trees that are robust to distribution shifts and formulated a two-stage robust MILP, solved via delayed constraint generation, iteratively separating worst-case scenarios.

\subsection{Audit and Verification}
\label{sec:adversarial}

Robustness assessment aims to determine whether adversarial examples exist, i.e., whether there is a perturbation $\boldsymbol{\delta} \in \mathcal{U}$ on a given input such that the model’s prediction changes or violates a prescribed network property. This problem naturally decomposes again into two dual tasks: \emph{auditing}, which searches for violations, and \emph{verification}, which aims to certify that no such violation exists within a prescribed perturbation set. Auditing methods aim to discover adversarial examples and can therefore provide concrete counterexamples when successful, but failure to find one does not constitute proof of robustness. Verification methods, in contrast, seek sound certificates that rule out the existence of adversarial examples, at the cost of significantly higher computational complexity. Despite substantial progress on both fronts, a large gap remains between the size and complexity of networks that can be fully verified and those deployed in practice, except in application domains where models are deliberately kept small and structured, e.g., safety-critical control systems in avionics \citep{Julian2016}. Recent empirical studies further indicate that no single verification algorithm consistently dominates across problem instances, suggesting complementarities between methods and motivating the development of hybrid approaches. We refer the reader to \citet{Konig2024} for a recent, more exhaustive methodological survey of the topic.

\paragraph{Auditing methods.}
Auditing approaches for neural networks primarily rely on heuristic search procedures to identify adversarial inputs. Deep neural networks are known to be particularly susceptible to adversarial perturbations \citep{Szegedy2013}, and most methods to find such perturbations are based on gradient information and local optimization, including fast single-step attacks such as FGSM \citep{Goodfellow2015}, multi-step projected gradient methods such as PGD \citep{Madry2018}, and stronger attacks such as those of \citet{Carlini2017}. Even LLMs are susceptible to jailbreaking (i.e., prompts designed to produce harmful content) via prompt-suffixes optimization through gradient search \citep{Zou2023}. These techniques, however, only provide empirical robustness assessments and do not offer guarantees: failure to find an adversarial example does not certify that none exists. Moreover, auditing methods can be fragile in poorly conditioned or highly nonconvex search spaces, leading to over-optimistic robustness assessments \citep{Uesato2018}. In settings where gradients are unavailable or unreliable, black-box and randomized strategies (e.g., decision-based or query-based attacks) provide complementary auditing tools \citep{Brendel2018}.

\paragraph{Compact models for neural-network verification.}
Several early studies established that feedforward networks with ReLU activations can be verified exactly using MILP formulations \citep{Lomuscio2017,Fischetti2018,Tjeng2019}. These approaches are both \emph{sound} and \emph{complete}: they never certify a false property and, given sufficient time, either produce a concrete counterexample or prove that none exists in the specified domain.
Subsequent work focused on strengthening these formulations by improving bounds on pre-activation variables. Tighter bounds yield stronger LP relaxations, enabling faster pruning in B\&B, and allowing earlier identification of \emph{stable} ReLU neurons whose activation state is fixed and whose binary variables can be eliminated. Notably, \citet{Anderson2020b} relied on an ideal convex-hull formulation for individual ReLU and max-pooling neurons over bounded domains. While this formulation yields the strongest possible single-neuron relaxation (see Highlight~\ref{highlight:relu-formulations}), instantiating it naively across all neurons introduces many additional variables and does not yield an ideal formulation at the network level due to interdependent activation patterns. As an alternative, \citet{Tsay2021} proposed partition-based formulations that group input variables and form the convex hull only over each group, yielding a hierarchy of intermediate relaxations between the classical big-$M$ model and the full convex hull. Beyond MILP, some works have also explored SDP relaxations \citep{Raghunathan2018,Dathathri2020}, which can better capture interactions between neurons but generally do not scale well with network size. Overall, MILP or SDP formulations rapidly attain their scalability limits on networks with a few thousand ReLU units, motivating extensive ongoing research towards alternative verification paradigms.

\begin{figure*}[!htb] \refstepcounter{highlight}
\centering \scalebox{0.95}{
\begin{highlight}{Alternative Formulations for a Single ReLU Neuron}
\label{highlight:relu-formulations}
\vspace*{0.3cm}

Consider a ReLU neuron in a layer, which computes
$$\hat{y} = \mathrm{ReLU}\!\left(\mathbf{w} \mathbf{x} + w_0\right)
      = \max\!\left\{0,\;\mathbf{w} \mathbf{x} + w_0\right\}.$$
Assume componentwise bounds on the input,
$\mathbf{x} \in [\mathbf{L},\mathbf{U}]$, and valid scalars
$$M^- \;\le\;
\min_{\mathbf{x}\in [\mathbf{L},\mathbf{U}]}
\left(\mathbf{w} \mathbf{x} + w_0\right),
\qquad
M^+ \;\ge\;
\max_{\mathbf{x}\in [\mathbf{L},\mathbf{U}]}
\left(\mathbf{w} \mathbf{x} + w_0\right).$$

\paragraph{(a) Big-$M$ formulation \citep{Fischetti2018,Tjeng2019}.}
The ReLU nonlinearity can be modeled using exactly one binary activation variable
$\eta$:
\begin{equation}
\left\{
\begin{aligned}
\hat{y} &\ge \mathbf{w} \mathbf{x} + w_0, \\
\hat{y} &\le \mathbf{w} \mathbf{x} + w_0 - M^-(1-\eta), \\
\hat{y} &\le M^+ \eta, \\
(\mathbf{x},\hat{y}) &\in [\mathbf{L},\mathbf{U}] \times \mathbb{R}^+, \\
\eta &\in \{0,1\}.
\end{aligned}
\right.
\label{eq:relu_bigM}
\end{equation}

\paragraph{(b) Extended formulation \citep{Anderson2020b}.}
A stronger formulation is obtained by modeling the active/inactive disjunction
in an extended variable space:
\begin{equation}
\left\{
\begin{aligned}
(\mathbf{x},\hat{y})
  &=  (\mathbf{x}^{-},\hat{y}^{-}) + (\mathbf{x}^{+},\hat{y}^{+}), \\
\hat{y}^{-} &= 0 \geq \mathbf{w} \mathbf{x}^{-} + w_0(1-\eta), \\
\hat{y}^{+} &= \mathbf{w} \mathbf{x}^{+} + w_0 \eta \geq 0, \\
\mathbf{L}(1-\eta)
  &\le \mathbf{x}^{-} \le \mathbf{U}(1-\eta), \\
\mathbf{L}\eta
  &\le \mathbf{x}^{+} \le \mathbf{U}\eta, \\
\eta &\in \{0,1\}.
\end{aligned}
\right.
\label{eq:relu_extended}
\end{equation}

Formulation~(a) is compact but admits weak relaxations when the bounds $M^-$ and $M^+$ are loose.
Formulation~(b) is locally ideal: relaxing $\eta \in [0,1]$ yields the convex hull for a single neuron. However, it introduces $O(\dim(\mathbf{x}))$ additional variables and constraints per neuron and does not yield an ideal formulation for the full network, as its LP relaxation can mix incompatible activation patterns among exponentially many possibilities. As a result, despite being theoretically stronger, (b) may perform worse in practice, motivating other intermediate formulations balancing compactness and relaxation strength \citep{Tsay2021}.
\end{highlight}}
\end{figure*}

\paragraph{Hybrid neural-network verification methods.}
State-of-the-art neural network verifiers no longer rely solely on compact MILP or SDP formulations. Instead, the dominant paradigm, adopted by all top-performing tools in recent editions of the International Verification of Neural Networks Competition (VNN-COMP; \citealt{Brix2024,Kaulen2025}), combines fast bound-propagation techniques with B\&B search.

A particularly successful family of methods is based on CROWN-style linear bound propagation. The original CROWN framework computes linear bounds on network outputs under input perturbations \citep{Xu2020}. These bounds were subsequently tightened by optimizing relaxation parameters via gradient ascent (the $\alpha$-CROWN approach; \citealt{Xu2021}), and integrated into a B\&B framework that explicitly branches on ReLU activation states (the $\beta$-CROWN component; \citealt{Wang2021c}). This combination yields verifiers that are both sound and complete while scaling effectively through GPU use. Extensions such as GenBaB generalize this framework to handle non-ReLU nonlinearities and general computational graphs \citep{Shi2025}.
To further strengthen relaxations, recent work incorporates additional constraints and cuts. General cutting-plane methods \citep{Zhang2022b,Zhou2024} leverage either MIP-generated cuts or cuts inferred during B\&B to capture multi-neuron interactions, while still relying on bound propagation for scalability. More recently, hybrid approaches such as SDP-CROWN inject semidefinite programming bounds into the bound-propagation pipeline, improving tightness while scaling to networks with tens of thousands of neurons \citep{Chiu2025}.

Alternative verification paradigms have also been developed alongside the $\alpha$-$\beta$-CROWN family. Some verifiers are grounded on SMT solving, such as Reluplex \citep{Katz2017} followed by Marabou \citep{Katz2019,Wu2024}, which performs logical reasoning over activation phases combined with linear feasibility checks. NeuralSAT adopts a DPLL(T)-style architecture inspired by SMT solving and achieved good performance (second position at VNN-COMP 2025) by combining Boolean reasoning over activation patterns with LP-based reasoning, abstraction, and neuron stabilization \citep{Duong2024b,Duong2024a}. Finally, reachability- and abstract-interpretation-based tools such as PyRAT compute sound over-approximations of reachable sets, which are then combined with splitting and refinement strategies.

\paragraph{Verification of other ML models.}
Robustness verification has also been studied for other predictive models. Dedicated approaches have been developed, among others, for graph neural networks \citep{Zhang2023,McDonald2024} and binarized neural networks \citep{Narodytska2018a,Khalil2019a,Jia2020}. For tree ensemble models, exact robustness and evasion analyses can be formulated as MILPs or SMTs \citep{Kantchelian2016,Chen2019a,Ahmad2025}. Robustness analysis has likewise been investigated for nonparametric models such as $k$-nearest neighbors, where verification exploits the geometric properties of Voronoi partitions \citep{Sitawarin2021}. Across different model families, the feasibility of verification depends heavily on the model itself: simpler, more structured hypothesis classes (such as shallow decision trees, small ensembles, or low-$k$ nearest neighbors) can greatly reduce combinatorial complexity and make exact certification practical at larger scales.

\paragraph{Verifiability by design.}
The approaches discussed so far treat verification as a post hoc analysis applied to a fixed trained model. A complementary line of work instead co-designs model architectures and training objectives to facilitate certification downstream. For instance, \citet{Xiao2019} introduced training procedures that explicitly promote ReLU stability, significantly reducing the number of activation patterns that exact verifiers need to consider. On a different research direction, \citet{Petersen2022,Petersen2024} showed how to train neural architectures composed exclusively of logical binary gates, and \citet{Kresse2025} highlighted that such networks are amenable to more efficient SAT-based formal verification at scale. More broadly, verifiability by design can be viewed as optimizing not only predictive performance under perturbations but also other structural properties that impact the difficulty of formal verification.

\subsection{Selective Classification and Rejection Learning}

While some models considered in Section~\ref{sec:training-robust} eliminate some outlier observations during training, they still aim to produce a prediction mechanism that covers all possible inputs. An alternative strategy is to allow models to \emph{abstain} or \emph{defer} predictions to humans on inputs deemed unreliable. This idea dates back to work on classification with a reject option \citep{Chow1970}, where optimal decision rules trade off misclassification error against rejection. Later formulations cast selective prediction as a joint optimization problem over a classifier and a rejector, introducing discrete decisions that determine whether to predict, abstain, or delegate the decision \citep{Cortes2016}. \citet{Mozannar2023} showed that learning optimal linear classifier-rejector pairs is NP-hard, and proposed MILP formulations that compute globally optimal deferral policies. These models are closely related to hybrid prediction models, as both create explicit routing or gating decisions that determine which component handles a given input, albeit optimized here for robustness rather than interpretability. Fairness considerations can also be enforced by adding explicit constraints on abstention, coverage, or group-wise error rates \citep[see][]{Mozannar2023,Yin2024}.

\subsection{Robust Counterfactual Explanations}

Robustness is a concern for predictive models but also for the explanations themselves.
Counterfactual explanations that are valid only for a specific trained model and that fail under small changes in the input, data, or model parameters can be misleading in deployment and undermine trust. This has motivated a growing line of work on robust counterfactual explanations, which aim to guarantee validity under retraining, with or without distribution shift, as well as query or algorithmic uncertainty.

Early work has highlighted the fragility of standard counterfactual explanations under \emph{predictive multiplicity}, that is, when different models can arise from randomized training or retraining procedures. In such settings, optimal counterfactuals often lie close to decision boundaries and are therefore particularly unstable. One approach to mitigating this issue explicitly models uncertainty in the learning process and enforces robustness requirements through chance constraints. Notably, \citet{Forel2024a} studied robust counterfactual explanations for random forests and ensemble models by treating explanation validity as a probabilistic guarantee over retraining randomness. Leveraging results from stochastic optimization, they derived a deterministic reformulation with robustness-aware feasibility thresholds.

A complementary line of work addresses robustness to uncertainty in future data distributions and in the models obtained after retraining. \citet{Bui2022,Bui2025} introduced counterfactual plans under distributional ambiguity, providing probabilistic coverage guarantees that a set of recourse actions remains valid across plausible future populations and retrained models. These approaches rely on distributionally robust optimization to balance proximity, diversity, and robustness. In contrast, \citet{Jiang2023a} and \citet{Jiang2024a} considered robustness under explicit uncertainty sets over model parameters, using interval abstractions and MILP formulations to produce robust explanations.

Beyond uncertainty in models or data, additional work has emphasized robustness in the execution of recourse itself. \citet{Maragno2024} proposed a robust optimization framework for computing \emph{regions} of counterfactual explanations that remain valid under small perturbations of the implemented actions. Rather than returning a single-point explanation, their approach computes a full region of feasible counterfactuals by solving a min–max robust optimization problem via iterative constraint generation.


\section{CO for ML Privacy}

A central question in privacy-preserving ML is whether sensitive information (e.g., about individuals, training data, or the learned model itself) can be inferred from released artifacts such as trained parameters, prediction APIs, gradients, or aggregate statistics. These privacy threats span a broad spectrum, including training data reconstruction \citep{Fredrikson2015}, membership inference \citep{Shokri2017}, and model extraction attacks \citep{Tramer2016}.

Many of these risks can be unified under an \emph{identifiability} perspective: given an observable artifact and partial knowledge of the learning or release mechanism, an adversary asks whether there exists a feasible or high-likelihood reconstruction, subject to known constraints, that reveals private information.
Depending on the threat model, such a reconstruction may correspond to an entire dataset, a subset of influential samples, the presence of a specific individual, or a high-fidelity surrogate of the model itself. Privacy is preserved only if the space of feasible reconstructions remains sufficiently large or uninformative.

\subsection{Privacy Auditing as Inverse Optimization}

This identifiability viewpoint naturally leads to \emph{privacy auditing} problems formulated as inverse optimization tasks.
In these settings, an attacker solves feasibility or optimization problems subject to structural constraints induced by model architectures, training dynamics, feature domains, encodings (e.g., one-hot constraints), or plausibility assumptions. This perspective also relates to database reconstruction attacks, in which adversaries solve optimization problems to recover private attributes from released aggregates~\citep{Dinur2003}. Such concerns played a fundamental role in motivating the deployment of differential privacy in large statistical agencies~\citep[see, e.g.,][]{Abowd2018}.

\subsubsection{Training-data leakage: Reconstruction and membership inference}

\paragraph{Dataset reconstruction from tree ensembles.} 
Tree-based models expose rich structural information through their leaves, class counts, and prediction probabilities. \citet{Ferry2024} demonstrated that such information can suffice, in realistic regimes, to reconstruct a training dataset exactly or near-exactly from white-box access to a trained random forest. They formalized dataset reconstruction as a CP model whose decision variables encode feature values, class labels, leaf assignments, and bagging multiplicities, while constraints enforce consistency with the tree paths and the published counts (see Highlight~\ref{highlight:DRAFT}). The high recovery rates observed in practice suggest that privacy auditing is well-posed as a combinatorial feasibility problem, going beyond statistical estimates based on mutual information calculations \citep[see, e.g.,][]{Bassily2018,Wang2021b}. Moreover, privacy threats are even more stringent if multiple models (e.g., from a Rashomon set) trained on the same data are released \citep{Hsu2025}, or if additional structural model constraints (e.g., to promote fairness) are publicly known \citep{Ferry2022}.

\begin{figure*}[!htb] \refstepcounter{highlight}
\centering \scalebox{0.95}{ \begin{highlight}[ycornermulticlass]{Dataset Reconstruction from Random Forests~\citep{Ferry2024}}
\label{highlight:DRAFT}
\begin{subequations}
\begin{align}
    \max_{\mathbf{z},\mathbf{\Tilde{x}},\mathbf{\Tilde{y}}, \mathbf{q}} \quad & \sum_{t \in \mathcal{T}} \sum_{i \in [n]} \sum_{o \in \mathcal{O}} \log (p_{o})q_{tio}\label{eq:draft_obj_bagging} \\
    \text{s.t.} \quad &  \sum_{y \in \mathcal{Y}} \Tilde{y}_{iy} = 1, & \forall i \in [n] \label{eq:draft_class_assignment_bagging} \\
    & \Tilde{y}_{iy} = 0 \Rightarrow \sum_{t \in \mathcal{T},~j\in \mathcal{L}_t} z_{tjiy} = 0 & \forall i \in [n], y \in \mathcal{Y} \label{eq:draft_class_coherence}\\
    & n_{tjy} = \sum_{i \in [n]} z_{tjiy}, & \forall t \in \mathcal{T}, j \in \mathcal{L}_t, y \in \mathcal{Y} \label{eq:draft_count_constraint_bagging} \\
    & \sum_{y \in \mathcal{Y}} z_{tjiy} \geq 1 \Rightarrow  \bigwedge_{(f, v) \in \Phi_j^{-}} \Tilde{x}_{if} \leq v & \forall i \in [n], t \in \mathcal{T}, j \in \mathcal{L}_t \label{eq:draft_consistency_bagging_1} \\
    & \sum_{y \in \mathcal{Y}} z_{tjiy} \geq 1 \Rightarrow \bigwedge_{(f, v) \in \Phi_j^{+}} \Tilde{x}_{if} > v & \forall i \in [n], t \in \mathcal{T}, j \in \mathcal{L}_t \label{eq:draft_consistency_bagging_2} \\
    & q_{tio} = 1 \quad \Leftrightarrow \sum_{j \in \mathcal{L}_t} z_{tjiy} = o & \forall t \in \mathcal{T}, i \in [n], o \in \mathcal{O} \label{eq:draft_bagging_link}
\end{align}
\end{subequations}

The reconstructed dataset is represented by variables $\Tilde{x}_{if}$ and $\Tilde{y}_{iy}$, which encode the value of feature $f$ for example $i$, and its membership to class $y$, respectively. Integer variable $z_{tjiy}$ denotes how many times example $i$ is assigned to leaf $j$ of tree $t$ with class $y$. Indeed, since each tree is trained on a bootstrapped version of the ensemble's training set, an example may appear zero, one, or multiple times in a given tree. Binary variable $q_{tio}$ indicates whether example $i$ appears exactly $o \in \mathcal{O}$ times in the training set of tree $t$, where $\mathcal{O}$ is chosen large enough to ensure feasibility with high probability.

The objective~\eqref{eq:draft_obj_bagging} maximizes the likelihood of the reconstructed dataset using the theoretical bootstrap probabilities $p_o$ that an example appears exactly $o$ times in the training set of a given tree. 
Constraints~\eqref{eq:draft_class_assignment_bagging} and~\eqref{eq:draft_class_coherence} ensure that each example is assigned to exactly one class and only contributes to counts for that class. Constraints~\eqref{eq:draft_count_constraint_bagging} link the observed leaf-wise class counts $n_{tjy}$ to these assignments. Constraints~\eqref{eq:draft_consistency_bagging_1} and~\eqref{eq:draft_consistency_bagging_2} enforce consistency between $\Tilde{\mathbf{x}}_i$ and the split conditions along the root-to-leaf path to leaf $j$, where $\Phi_j^{-}$ (respectively, $\Phi_j^{+}$) denotes the set of split conditions (attribute-value tuples) for which the path branches left (respectively, right). Finally, Constraints~\eqref{eq:draft_bagging_link} one-hot encode the number of times each example appears in each tree’s bootstrap sample. 
\end{highlight}} \end{figure*}

An extension of this audit paradigm involves studying \emph{differentially private} forests, in which leaf counts are perturbed (e.g., via Laplace noise). Since several differentially private training mechanisms explicitly perturb leaf statistics or splitting decisions \citep{Fletcher2020}, reconstruction becomes a noisy inverse problem in which both the underlying dataset and the unknown noise realizations must be inferred. \citet{Gorge2026} cast this task as a likelihood-maximization CP model consistent with the forest structure and the differential privacy mechanism. Empirically, they showed that while meaningful differential privacy guarantees reduce reconstruction success, privacy leaks are fully avoided only when predictive utility becomes nearly trivial.

\paragraph{Reconstruction from neural networks.}
For neural networks trained with gradient-based methods, training-data leakage can also be cast as a structured inverse optimization problem: the attacker searches for inputs whose optimality conditions are consistent with the released parameters. For homogeneous networks trained with exponentially-tailed losses, implicit-bias results show that gradient flow (and, empirically, gradient descent) converges in direction toward solutions satisfying the Karush-Kuhn-Tucker (KKT) conditions of a max-margin problem. Exploiting this structure, \citet{Haim2022} formulated reconstruction as minimizing violations of KKT conditions over candidate samples and dual variables, yielding a nonconvex optimization problem whose decision variables include both reconstructed inputs and Lagrange multipliers. \citet{Buzaglo2024} extended this approach to multiclass settings and to broader loss functions, notably by incorporating weight decay, and empirically studied how regularization choices and model size affect reconstructability. More recent work by \citet{Oz2024} further evaluated these KKT-based reconstruction methods on real-world models trained via transfer learning.
Overall, these works suggest that leakage within neural networks is concentrated on margin samples, but architectural and optimization choices can significantly affect the attack surface.

\paragraph{Gradient-based reconstruction and federated learning.} In distributed and federated learning, gradients or model updates are often treated as privacy-preserving proxies for raw data, with only these updates being shared across participants. However, gradient inversion attacks demonstrate that individual data points can be reconstructed from gradients in various realistic settings, including deep architectures and aggregated updates \citep{Geiping2020,Lei2019,Wang2023}. From a CO viewpoint, gradient inversion can be viewed as a constrained inverse problem: the attacker seeks inputs (and sometimes labels) that match the observed gradients under domain constraints. While the general problem is nonconvex and can be NP-hard, several tractable subcases exist (e.g., analytic inversion for certain fully connected layers, or LP-based inversion for specific ReLU generative structures), and modern solvers or optimization methods can succeed at scale \citep{Lei2019, Geiping2020}. More recent work further shows that \emph{active} adversaries can amplify leakage by manipulating the protocol (e.g., by sending slightly modified weights/queries), transforming accidental leakage into systematic extraction \citep{Boenisch2023}.

\paragraph{Reconstruction of time series.}
Beyond attacks against predictive models, recent work also shows that reconstruction attacks extend to analytical data structures released for time series analysis. Matrix profiles (MPs) are widely used summaries of time series, and are often viewed as privacy-preserving and safe to share with analysts. However, \citet{Zhang2026} demonstrated that MPs can enable accurate reconstruction of the original time series. They cast the reconstruction task as a constraint-optimization problem induced by the MP structure and considered both a discretized variant tackled with CP and a main approach based on continuous nonlinear optimization over real-valued signal variables. The resulting reconstructions can then be refined using domain-specific knowledge, such as known ECG waveform patterns.

\subsubsection{Model leakage: Model extraction attacks}

Privacy threats in ML also concern model confidentiality: an adversary may seek to reconstruct the prediction function itself (or a high-fidelity surrogate), either for intellectual property theft or as a stepping stone for other attacks. Seminal work by \citet{Tramer2016} showed that many deployed machine learning as a service (MLaaS) APIs leak enough information (e.g., confidence values) to enable efficient extraction of popular model classes, including linear models and decision trees, using only black-box queries. Subsequent work connected model extraction to active learning, framing extraction as an adaptive query-design problem in which each request reveals additional information about decision boundaries \citep{Chandrasekaran2020}.

A recent line of work analyzes model extraction from counterfactual explanation queries through the lens of \emph{competitive analysis}, i.e., comparing an online querying strategy to an optimal offline strategy with full knowledge of the target \citep{Khouna2025}. In this framework, the extraction attack is viewed as an online discovery problem: each oracle query reveals partial information about the hypothesis space, and the goal is to minimize the number of queries required to recover a functionally identical model under worst-case conditions. The study focuses on additive decision tree models (decision trees, random forests, boosted trees) and derives worst-case competitive-ratio guarantees, as well as proposing reconstruction algorithms with strong anytime behavior.

Moreover, \citet{Otto2026} investigated extraction attacks against linear classifiers using counterfactual and robust counterfactual queries. Leveraging the geometric structure of linear decision boundaries, they established upper bounds on the number of queries required to recover the model parameters. Remarkably, when the counterfactual distance is differentiable (e.g., the $\ell_2$ norm), a single counterfactual query suffices to recover the full parameter vector. In contrast, for non-differentiable norms such as $\ell_1$ or~$\ell_\infty$, $p+1$ independent counterfactual queries are required. For robust counterfactuals, additional prediction queries become necessary.

\subsection{Protecting Private Information}

\paragraph{Statistical disclosure control.}
Long before modern ML, the statistical disclosure control literature studied various optimization formulations to prevent inference from aggregate data released to the public \citep{Castro2012}. A canonical example is the \emph{cell suppression problem} (CSP), where the goal is to suppress a minimum-cost set of table entries such that no sensitive cell can be reconstructed from released row and column totals. Early work showed that CSP is strongly NP-hard, although the attacker’s reconstruction problems admit a network-flow representation in the special case of two-dimensional tables, enabling exact min-cost flow and cut-based algorithms for privacy audits and optimal suppression \citep{Kelly1992,Fischetti1999}. For higher-dimensional, hierarchical, or linked tables, this structure no longer applies, requiring more general MILP formulations and Benders-based decomposition methods \citep{Fischetti2001,Baena2020}. Beyond pure suppression, some work has also exploited continuous-discrete optimization models for controlled tabular adjustment, which replaces sensitive values by minimally distorted alternatives \citep{Glover2008}.

\paragraph{Privacy-preserving learning and model selection.}
CO methods have also been harnessed to help construct differentially private ML models.
A first line of work employs optimization solely as a postprocessing step for differentially private releases, leveraging the fact that differential privacy guarantees are preserved under deterministic, data-independent postprocessing. \citet{Hay2010} and \citet{Qardaji2013} showed that noisy tabular and histogram releases can be substantially improved by solving constrained optimization problems that enforce structural consistency, such as non-negativity, additivity, and hierarchical relations. Building on this idea, \citet{Fioretto2019} and \citet{Fioretto2021} developed optimization-based frameworks for the private release of time-series and hierarchical census data. Their methods combine private sampling and perturbation with constrained optimization and dynamic programming to respect problem structure (e.g., temporal smoothness or hierarchical aggregation) while redistributing noisy estimates.

More recently, \citet{Prastakos2025} studied high-dimensional sparse variable selection under pure differential privacy, where releasing a feature subset should not allow an adversary to infer whether any particular data point is present in the dataset. To avoid the computational cost of applying the exponential mechanism over an exponentially large set of feature subsets, they used MIP to identify the best and near-best feature subsets. Sampling from this reduced set then yields scalable algorithms with pure differential privacy guarantees and strong empirical performance compared to convex relaxations or Markov chain Monte Carlo (MCMC) methods.

\paragraph{Data minimization during inference.}
Training-time data consumption and protection are not the only concerns regarding privacy breaches. In practice, there is also growing interest in minimizing data consumption during inference. From this perspective, models differ markedly in the amount of information they inherently require to produce a prediction. Decision trees, for example, only query feature values along their path to a leaf. Similarly, in rule lists and rule sets, inference stops as soon as a single rule is satisfied. In contrast, most neural networks and linear models assume that all input features are available at prediction time, making inference intrinsically data-hungry.

Early work on \emph{active learning} \citep{Settles2009} and cost-sensitive \emph{active classifiers} \citep{Greiner2002} already formalized inference as a sequential optimization problem, in which the classifier adaptively selects which feature tests to acquire in order to minimize expected misclassification loss under explicit acquisition costs. While this literature was primarily motivated by reducing labeling or measurement costs during learning, it naturally applies at test time when features correspond to sensitive attributes. In this framework, classifiers are viewed as policies over partially observed instances, and decision trees arise as solutions to a stochastic dynamic program over feature subsets. Two decades later, inference-time data consumption has become a first-class concern for regulatory compliance, as discussed in \citet{Staab2025}. Extending beyond inherently sparse models, \citet{Tran2023} proposed a formal optimization framework for inference-time data minimization that, for a fixed trained predictor, identifies a minimal subset of features sufficient to certify the prediction under worst-case completions of the remaining attributes. This recasts data exposure minimization as a sequence of verification problems over different feature subsets.

\section{Discussions and Research Perspectives}
\label{sec:conclusion}

As seen in this survey, CO tools can play a central role across a wide range of trustworthy machine learning (ML) tasks, including training, simplification, explanation, fairness and robustness auditing, and privacy auditing or protection. Through an analysis of over three hundred contributions, we have seen how mathematical models and formal verification methods offer capabilities not obtainable with purely heuristic or gradient-based approaches, progressively shaping an interdisciplinary research field at the crossroads of CO and ML.

\paragraph{Solver improvements reshape solution capabilities.}
A key takeaway is the dramatic progress in both specialized CO solvers and applications that harness them for trustworthy ML over the past decades. Advances in MILP, CP, SAT, SMT, and other related solvers have steadily expanded the scale of problems that can be addressed \citep{Koch2022,Clautiaux2025}. Consequently, approaches initially viewed as primarily theoretical become tractable for instances of practical relevance, calling for regular reassessment of the role of alternative CO techniques across the different areas of trustworthy ML. At the same time, progress has been uneven across areas: for example, while extensive work has explored sparse and robust training, far fewer studies have leveraged CO for privacy-preserving ML.

\paragraph{Global optimization and certification are indispensable for trust.}
In many high-stakes applications, stakeholders ultimately want assurances that certain behaviors cannot occur, particularly in settings subject to safety or regulatory constraints. At the same time, as regulatory frameworks continue to evolve, concrete operational definitions of concepts such as explainability, transparency, and privacy are still often lacking, largely because establishing such guarantees for modern deep learning models remains a formidable challenge. As a consequence, there is a wide gap between our abilities to train increasingly complex systems and our abilities to formally verify that these systems behave as intended.

In view of this, mathematically grounded optimization and formal verification methods offer a path toward meaningful guarantees that avoid the inconsistent, unfair, or infeasible outcomes associated with heuristic approaches. At the same time, their applicability remains focused on models that are sufficiently simple and structured to admit formal analysis. This raises a broader question: should regulatory standards and deployment practices distinguish application domains where formal certification is essential and those where it is not, and (if so) should this implicitly favor model classes that are more interpretable and auditable, even when more complex alternatives achieve slightly higher predictive accuracy?

\paragraph{Models need to be designed for certifiability.}
Related to the previous point, an important research direction is the co-design of models that consider downstream certifiability and explainability requirements. When several models achieve comparable predictive performance, selecting those that are structurally favorable (e.g., sparser or more stable) can dramatically improve the tractability of verification methods. From this perspective, verification must not be viewed solely as a post hoc analysis, but also as a criterion guiding model choice.
Moreover, model choice is not always limited to training from scratch. In many cases, complex predictors can be transformed into alternative representations (sometimes even faithfully, without altering their prediction function) such as decision trees, rule sets, or other structured surrogates. Different representations of the same underlying predictor yield very different explainability, robustness, or verification properties. Additionally, hybrid models \citep{Ferry2024a} offer another promising avenue by combining a simple, interpretable component that handles most inputs with a more complex fallback model for rare or ambiguous cases. Such designs suggest that certifiability and interpretability can be optimized where they matter most.

\paragraph{An optimization mindset clarifies benchmarking and evaluation.} Closely related to the question of certification is how trustworthy ML methods are evaluated. Many works simultaneously introduce new tasks, models, and algorithms, yet assess performance using task-specific metrics without clearly measuring success on the underlying optimization problem. In contrast, CO-based formulations make objectives, constraints, and trade-offs explicit, naturally leading to evaluations in terms of optimality gaps, certificates, or Pareto frontiers. While such objective-driven evaluations may appear more abstract, they are in fact complementary to empirical metrics: they help disentangle limitations of the model itself from those of the solution method, and avoid conflating heuristic convergence issues with fundamental inadequacies of a given model class \citep{Le2025a}.

\paragraph{Scalability hinges on exploiting the full spectrum of CO techniques.}
Scalability rarely comes from a single modeling paradigm. Tailored B\&B algorithms, dynamic programming with branch-and-bound (DPB\&B), problem-specific relaxations, and hybrid search strategies outperform generic MILP or SAT approaches across many applications. Consequently, the choice among MILP, CP, SAT, SMT, or hybrid approaches is fundamental and requires careful evaluation, as each formalism offers distinct capabilities. Specialized libraries are emerging to facilitate the integration of ML models into optimization pipelines (see, e.g., \url{https://github.com/Gurobi/gurobi-machinelearning}). Conversely, mature CO-oriented libraries for trustworthy ML tasks are still largely missing.

\paragraph{Trustworthy ML requires navigating inherent trade-offs.}
Throughout the survey, we observed persistent tensions between different trustworthiness desiderata such as accuracy, robustness, fairness, interpretability, and privacy. These objectives are rarely aligned, and improving one often degrades another \citep{Ferry2023a,Langlade2025}, whereas techniques designed to provide users with stronger accountability or explainability guarantees can, in some cases, increase exposure of models or data and thereby jeopardize privacy \citep{Khouna2025}. CO tools are particularly well-suited for navigating such tensions, as they allow multiple criteria to be modeled jointly and parameterized explicitly. By enabling systematic exploration of Rashomon sets of good models, optimization-based approaches reveal achievable trade-offs and support more principled design choices, in contrast to heuristics that often obscure these relationships.

\paragraph{Guarantees must ultimately be assessed at the decision level.}
Finally, throughout this survey, trustworthy ML has largely been examined through a predictive lens. In practice, however, ML models are often embedded within larger decision-making pipelines, as in end-to-end learning and contextual optimization \citep{Sadana2025}. In various application domains that combine CO and ML, such as automated trading, pricing, resource allocation, demand forecasting, or real-time decisions in transportation and logistics, actions are executed rapidly with limited human oversight. In such settings, predictive accuracy alone is not the relevant metric: what ultimately matters is the downstream impact of prediction errors on decisions, costs, fairness outcomes, and safety risks.

This perspective calls for a broader notion of \emph{trustworthy decision-focused learning}, in which robustness, fairness, and accountability are assessed at the decision level rather than the prediction level. Classical notions of trustworthiness developed in the ML literature should be revisited to account for these challenges (e.g., as done in \citealt{Forel2023}). Moreover, decision-making is increasingly delegated to multiple interacting autonomous agents, each relying on learned components and responding to others’ actions. These strategic interactions introduce feedback loops that cannot be captured by single-model analysis. Ensuring guarantees in such systems requires integrating learning, optimization, and game-theoretic reasoning to analyze stability, incentives, and robustness in complex decision environments.\\

In summary, CO provides a language and toolkit for addressing many challenging aspects of trustworthy ML. While current methods face scalability limits, depending on the model and application, the rapid evolution of solvers, hybrid algorithms, and problem formulations suggests that the boundary of what can be certified will continue to expand. Bridging optimization, learning, and decision-making in a principled and scalable manner is both a central challenge and a major opportunity for future research.

\section*{Acknowledgments}
This research was supported by the Fonds de recherche du Québec -- Nature et technologies (FRQNT) through the Team Research Project program, and by SCALE-AI through its Research Chairs program. The authors gratefully acknowledge this support. They also thank Pierre-Yves Bouchet and Rahul Mazumder for insightful discussions and valuable comments on earlier versions of this manuscript.

\appendix
\newpage
\section{Notations}\label{appendix:notations}

\begin{table}[h!]
\centering
\caption{Main notations used throughout this paper}\label{tab:notations_summary}
\vspace{5pt}
\resizebox{0.95\textwidth}{!}{
\begin{tabular}{l@{\hspace*{0.8cm}}l@{\hspace*{0.8cm}}p{9cm}}
\toprule
\textbf{Data/Predictors} & $n$               & Number of training examples                                                                                     \\
&$p$               & Number of dataset features                                                                                      \\
&$X$               & Training examples' features (matrix of size $n \times p$)                                                       \\
&$\mathbf{x}_i$    & Features' vector for example $i \in [n]$                                                                        \\
&$x_{if}$          & Value of feature $f \in [p]$ for example $i \in [n]$                                                            \\
&$\mathcal{X}$     & Input domain (s.t., $\mathbf{x} \in \mathcal{X}$)                                                               \\
&$\mathbf{y}$      & Training examples' labels (vector of size $n$)                                                                  \\
&$y_i$             & Ground truth label for example $i \in [n]$                                                                      \\
&$\mathcal{Y}$     & Output domain (s.t., $y \in \mathcal{Y}$) \\
&$h$ (or $g$)      & Predictive model s.t. $h: \mathcal{X} \mapsto \mathcal{Y}$                                                      \\
\midrule
\textbf{Variables} & $\mathbf{w}$      & Model parameters                                                                                                \\
&$w_0$             & Model parameter: bias/intercept                                                                                  \\
&$\xi_i$           & Loss variable for example $i \in [n]$                                                                \\
&$u$               & Typically a binary variable indicating whether a given token (e.g., rule, parameter, node) is used in the model \\
&$z$               & Typically a flow variable                                                                                       \\
\midrule
\textbf{Hyper-parameters}&$\lambda$         & Regularization coefficient, controlling the trade-off between utility and other desiderata                      \\
&$C$               & Coefficient controlling the trade-off between utility and other desiderata                                      \\
&$\gamma$          & Sparsity parameter: maximum number of non-zero parameters of a model                                            \\
&$ \epsilon$       & Small constant                                                                                                  \\
\midrule
\textbf{Others}& $[n]$             & For any integer $n$, denotes the set $\{1, \ldots, n\}$                                                         \\ 
\bottomrule
\end{tabular}
}
\end{table}

\end{document}